\documentclass[conference]{IEEEtran}
\IEEEoverridecommandlockouts
\usepackage{cite}
\usepackage{amsmath,amssymb,amsfonts}
\usepackage{algorithmic}
\usepackage{textcomp}
\usepackage{xcolor}
\usepackage{microtype}
\usepackage{graphicx}
\usepackage{booktabs} 
\usepackage{times}
\usepackage{epsfig}
\usepackage{hyperref}
\usepackage{shortcuts_js}
\usepackage[ruled,linesnumbered]{algorithm2e}
\usepackage{siunitx} 
\usepackage{multirow}
\usepackage{subfig}
\let\oldemptyset\emptyset

\usepackage{cleveref} 

\def\BibTeX{{\rm B\kern-.05em{\sc i\kern-.025em b}\kern-.08em
    T\kern-.1667em\lower.7ex\hbox{E}\kern-.125emX}}
\begin{document}

\title{Practical Design Space Exploration}


\author{\IEEEauthorblockN{1\textsuperscript{st} Luigi Nardi}
\IEEEauthorblockA{
\textit{Stanford University}\\
lnardi@stanford.edu}
\and
\IEEEauthorblockN{2\textsuperscript{nd} David Koeplinger}
\IEEEauthorblockA{
\textit{Stanford University}\\
dkoeplin@stanford.edu}
\and
\IEEEauthorblockN{3\textsuperscript{rd} Kunle Olukotun}
\IEEEauthorblockA{
\textit{Stanford University}\\
kunle@stanford.edu}
}

\maketitle

\begin{abstract}
Multi-objective optimization is a crucial matter in computer systems design space exploration because real-world applications often rely on a trade-off between several objectives.
Derivatives are usually not available or impractical to compute and 
the feasibility of an experiment can not always be determined in advance. 
These problems are particularly difficult when the feasible region is relatively small, and it may be prohibitive to even find a feasible experiment, let alone an optimal one.

We introduce a new methodology and corresponding software framework, HyperMapper 2.0, which handles multi-objective optimization, unknown feasibility constraints, and categorical/ordinal variables. This new methodology also supports injection of the user prior knowledge in the search when available. 
All of these features are common requirements in computer systems but rarely exposed in existing design space exploration systems.
The proposed methodology follows a white-box model which is simple to understand and interpret (unlike, for example, neural networks) and can be used by the user to better understand the results of the automatic search.


We apply and evaluate the new methodology to the  automatic static tuning of hardware accelerators within the recently introduced Spatial programming language, with minimization of design run-time 
and compute logic under the constraint of the design fitting in a target field-programmable gate array chip.
Our results show that HyperMapper 2.0 provides better Pareto fronts compared to state-of-the-art baselines, with better or competitive hypervolume indicator and with 8x improvement in sampling budget for most of the benchmarks explored.

\end{abstract}





\section{Introduction}
\label{introduction}

Design problems are ubiquitous in scientific and industrial achievements.
Scientists design experiments to gain insights into physical and social phenomena, and engineers design machines to execute tasks more efficiently.
These design problems are fraught with choices which are often complex and
high-dimensional and which include interactions that make them difficult for individuals to reason about.
In software/hardware co-design, for example, companies develop libraries with tens or hundreds of free choices and parameters that interact in complex ways.
In fact, the level of complexity is often so high that it becomes impossible to find domain experts capable of tuning these libraries \cite{conn2009introduction}.

Typically, a human developer that wishes to tune a computer system will try some of the options and get an insight of the response surface of the software.
They will start to fit a model in their head of how the software responds to the different choices.
However, fitting a complex multi-objective function without the use of an automated system is a daunting task. 
When the response surface is complex, e.g. non-linear, non-convex, discontinuous, or multi-modal,
a human designer will hardly be able to model this complex process, ultimately missing the opportunity of delivering high performance products. 

Mathematically, in the mono-objective formulation, we consider the problem of finding a global minimizer of an unknown (black-box) objective function $f$:
\begin{equation}\label{eq:optim_pb}
	x^* = \argmin_{x \in \bbX} f(x)
\end{equation}
where $\bbX$ is some input decision space of interest (also called design space).
The problem addressed in this paper is the optimization of a deterministic
function $f:\bbX \rightarrow \bbR$ over a domain of interest that includes lower and upper bound constraints on the problem variables.

When optimizing a smooth function, it is well known that 
useful information is contained in the function gradient/derivatives which can be leveraged, for instance, by first order methods. 
The derivatives are often computed by hand-coding, by automatic differentiation, or by finite differences.
However, there are situations where such first-order information is not available or even not well defined.
Typically, this is the case for computer systems workloads that include many discrete variables, \ie either categorical (\eg boolean) or ordinal (\eg choice of cache sizes), over which derivatives cannot even be defined.
Hence, we assume in our applications of interest that the derivative of $f$ is neither symbolically nor numerically available.
This problem is referred to in the literature as DFO \cite{conn2009introduction,rios2013derivative},
also known as black-box optimization \cite{feliot2017bayesian} and, in the computer systems community, as design space exploration (DSE) \cite{ipek2006efficiently,kang2010approach}.

\begin{table}[th]
    \begin{center}   
      \begin{tabular}{ | l | c | c | c | c |}
    \hline
    Name & Multi & RIOC var. & Constr. & Prior\\ \hline
    \hline
    GpyOpt & \nomark & \nomark & \nomark & \nomark\\ \hline
    OpenTuner & \nomark & \yesmark & \nomark& \nomark\\ \hline
    SURF & \nomark & \yesmark & \nomark & \nomark\\ \hline
    SMAC & \nomark & \yesmark  & \nomark & \nomark\\ \hline
    Spearmint & \nomark & \nomark  & \yesmark & \nomark\\ \hline
    Hyperopt & \nomark & \yesmark  & \nomark & \yesmark\\ \hline
    Hyperband & \nomark & \yesmark & \nomark & \nomark\\ \hline
    GPflowOpt & \yesmark & \nomark  & \yesmark & \nomark\\ \hline
    cBO & \nomark & \nomark & \yesmark & \nomark\\ \hline
    BOHB & \nomark & \yesmark & \nomark & \nomark\\ \hline
    HyperMapper 1.0 & \yesmark & \yesmark & \nomark & \nomark \\ \hline
    \textbf{HyperMapper 2.0} & \yesmark & \yesmark & \yesmark & \yesmark \\ \hline
      \end{tabular}
    \end{center}
  \caption{Derivative-free optimization software taxonomy.
  \textit{Multi} notes if the software is multi-objective or not.
  \textit{RIOC var.} says if the software supports all Real, Int, Ordinal and Categorical variables.
  \textit{Constr.} refers to inequality constraints that define a feasible region that are used in the optimization process. 
  \textit{Prior} represents the ability of the software to inject prior knowledge in the search.
  }
  \label{table_taxonomy_software}
\end{table}

In addition to objective function evaluations, 
many optimization programs have similarly expensive evaluations of constraint functions. The set of points where such constraints are satisfied is referred to as the \emph{feasibility} set.
For example, in computer micro-architecture, fine-tuning the particular specifications of a CPU (\eg L1-Cache size, branch predictor range, and cycle time)
need to be carefully balanced to optimize CPU speed, while keeping the power usage strictly within a pre-specified budget. 
A similar example is in creating hardware designs for field-programmable gate arrays (FPGAs). 
FPGAs are a type of reconfigurable logic chip with a fixed number of units available to implement circuits. Any generated design must keep the number of units strictly below this resource budget to be implementable on the FPGA. 
In these examples, feasibility of an experiment cannot be checked prior to termination of the experiment; this is often referred to as \emph{unknown feasibility} in the literature \cite{gelbart2014bayesian}.
Also note that the smaller the feasible region, the harder it is to check if an experiment is feasible (and even more costly to check optimality \cite{gelbart2014bayesian}).  


While the growing demand for sophisticated DFO methods has triggered the development of a wide range of approaches and frameworks, none to date are featured enough to fully address the complexities of design space exploration and optimization in the computer systems domain.
To address this problem, we introduce a new methodology and a framework dubbed HyperMapper 2.0.
HyperMapper 2.0 is designed for the computer systems community and can handle design spaces consisting of multiple objectives and categorical/ordinal variables. Emphasis is on exploiting user prior knowledge via modeling of the design space parameters distributions. Given the years of hand-tuning experience in optimizing hardware, designers bear a high level of confidence. HyperMapper 2.0 gives means to inject  knowledge in the search algorithm. This is achieved by introducing for the first time the use of a Beta distribution for modeling the user belief, i.e., prior knowledge, on how each parameter of the design space influences a response surface. 
In addition, bearing in mind the feasibility constraints that are common in computer systems workloads we introduce for the first time a model that considers unknown constraints, which is, constraints that are only known after evaluating a  system configuration. To aid comparison, we provide a list of existing tools and the corresponding taxonomy in \Cref{table_taxonomy_software}.
Our framework uses a model-based algorithm, i.e., construct and utilize a surrogate model of $f$ to guide the search process. A key advantage of having a model, and more specifically a white-box model, is that the final surrogate of $f$ can be analyzed by the user to understand the space and learn fundamental properties of the application of interest. 

As shown in Table~\ref{table_taxonomy_software}, HyperMapper 2.0 is the only framework to provide all the features needed for a practical design space exploration software in computer systems applications. 
The contributions of this paper are:
\begin{itemize}
\item A methodology for multi-objective optimization that deals with categorical and ordinal variables, unknown constraints, and exploitation of the user prior knowledge. 
\item An integration and experimental results of our methodology in a full, production-level compiler toolchain for hardware accelerator design.
\item A framework dubbed HyperMapper 2.0 implementing the newly introduced methodology, designed to be simple, user-friendly and application independent.
\end{itemize}

The remainder of this paper is organized as follows: 
\Cref{background} provides the problem statement and background. In \Cref{methodology}, we describe our methodology and framework. In \Cref{evaluation} we present our experimental evaluation. \Cref{related_work} discusses related work. We conclude in \Cref{conclusions} with a brief discussion of future work.


\section{Background}
\label{background}
In this section, we provide the notation and basic concepts used in the paper. We describe the mathematical formulation of the mono-objective optimization problem with feasibility constraints. We then expand this to a definition of the multi-objective optimization problem and provide background on randomized decision forests~\cite{breiman2001random}.

\subsection{Mono-objective Optimization with Unknown Feasibility Constraints}
Mathematically, in the mono-objective formulation, we consider the problem of finding a global minimizer (or maximizer) of an unknown (black-box) objective function $f$ under a set of constraint functions $c_i$:
\begin{equation*}
\begin{aligned}
& x^* = \argmin_{x \in \bbX}
& & f(x) \\
& \text{subject to}
& & c_i(x) \leq b_i, \; i = 1, \ldots, q,
\end{aligned}
\end{equation*}
where $\bbX$ is some input design space of interest and $c_i$ are $q$ unknown constraint functions.
The problem addressed in this paper is the optimization of a deterministic
function $f:\bbX \rightarrow \bbR$ over a domain of interest that includes lower and upper bounds on the problem variables.

The variables defining the space $\bbX$ can be real (continuous), integer, ordinal, and categorical. 
Ordinal parameters have a domain of a finite set of values which are either integer and/or real values. For example, the sets $\{1, 5, 8\}$ and $\{3.4, 2.5, 6, 9.1\}$ are possible domains of ordinal parameters.
Ordinal values must have an ordering by the less-than operator. 
Ordinal and integer cases are also referred to as discrete variables. 
Categorical parameters also have domains of a finite set of values but have no ordering requirement. For example, sets of strings describing some property like $\{true, false\}$ and $\{car, truck, motorbike\}$ are categorical domains.
The primary benefit of encoding a variable as an ordinal is that it can allow better inferences about unseen parameter values. With a categorical parameter, the knowledge of one value does not tell one much about other values, whereas with an ordinal value we would expect closer values
(with respect to the ordering) to be more related.

We assume that the derivative of $f$ is not available,
and that bounds, such as Lipschitz constants, for the derivative of $f$ is also unavailable.
Evaluating feasibility is often in the same order of expense as evaluating the objective function $f$. 
As for the objective function, no particular assumptions are made on the constraint functions.

\subsection{Multi-Objective Optimization: Problem Statement}
\label{problem_statement_multi}
\begin{figure}[tb]
\centering
\includegraphics[width=0.99\linewidth]{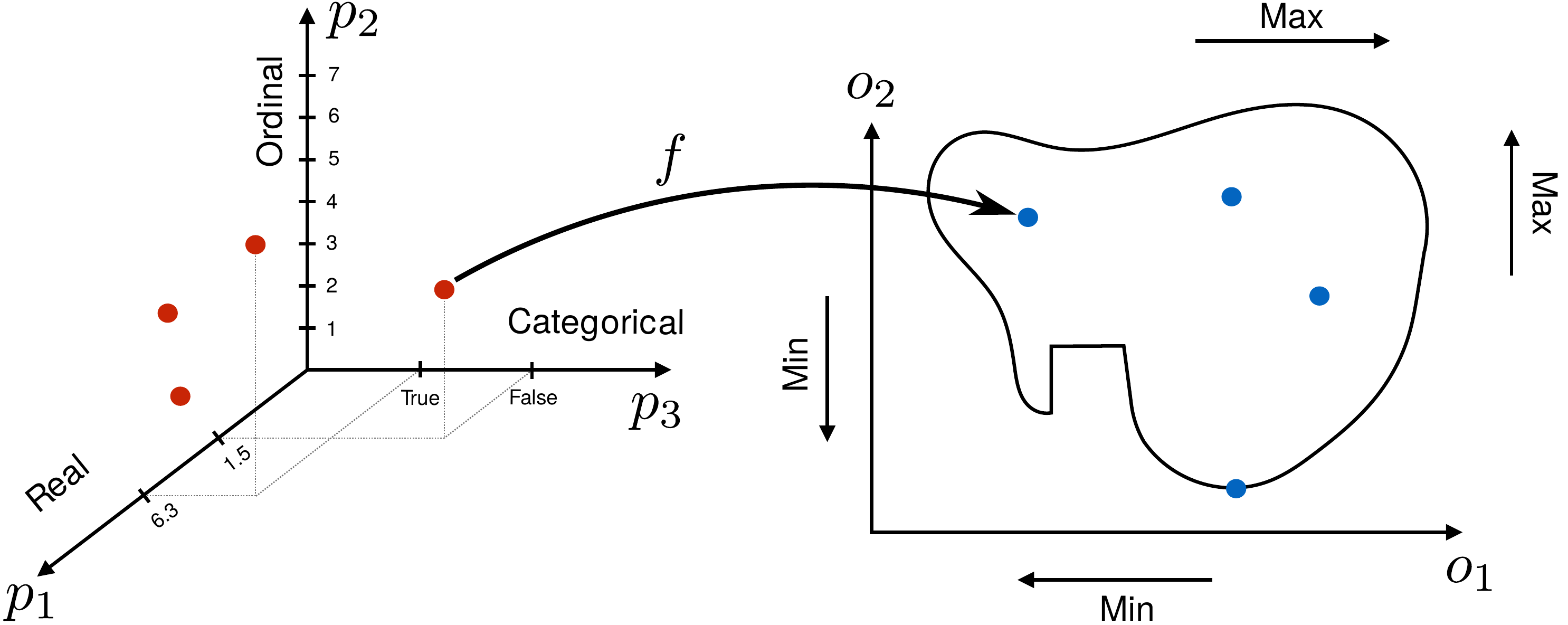}
\caption{Example of a multi-objective space. The multi-objective function $f$ maps each point in the 3-dimensional design space on the left to the optimization space on the right. }
\label{figure_multiobjective}
\end{figure}

A pictorial representation of a multi-objective problem is shown in Figure~\ref{figure_multiobjective}. 
On the left, a three-dimensional design space is composed by one ordinal ($p_1$), one real ($p_2$), and one categorical ($p_3$) variable. 
The red dots represent samples from this search space. 
The multi-objective function $f$ maps this input space to the output space on the right, also called the optimization space. 
The optimization space is composed by two optimization objectives ($o_1$ and $o_2$). 
The blue dots correspond to the red dots in the left via the application of $f$. 
The arrows Min and Max represent the fact that we can minimize or maximize the objectives as a function of the application. 
Optimization will drive the search of optima points towards either the Min or Max of the right plot. 

Formally, let us consider a multi-objective optimization (minimization) over a design space $\bbX \subseteq \mathbb{R}^{d}$.
We define $f:\bbX \rightarrow \mathbb{R}^{p}$ as our vector of objective functions $f = (f_{1}, \dots, f_{p})$,
taking $x$ as input, and evaluating $y = f(x)$. Our goal is to identify the Pareto frontier of $f$; that is,
the set $\Gamma \subseteq \bbX$ of points which are not dominated by any other point,
\ie the maximally desirable $x$ which cannot be optimized further for any single objective without making a trade-off.
Formally, we consider the partial order in $\mathbb{R}^{p}$: $y \prec y'$ iff $\forall i \in [p], y_{i} \leqslant y'_{i}$
and $\exists j, \, y_{j} \! < \! y'_{j}$, and define the induced order on $\bbX$: $x \prec x'$ iff $f(x) \prec f(x')$.
The set of minimal points in this order is the Pareto-optimal set $\Gamma = \{x \in \bbX : \nexists x'$ such that $x' \prec x\}$.

We can then introduce a set of inequality constraints $c(x) = (c_{1}(x), \dots, c_{q}(x))$, $b = (b_{1}, \dots, b_{q})$ to the optimization,
such that we only consider points where all constraints are satisfied ($c_{i}(x)\leqslant b_{i}$). These constraints directly correspond to real-world limitations of the design space under consideration.
Applying these constraints gives the constrained Pareto 
\begin{equation*}
\begin{aligned}
& \Gamma = \{ x \in \bbX : \forall i \leqslant q, \, c_{i}(x) \leqslant b_{i} \}\\
& \text{where}~\nexists x' \in \bbX~\text{such that}~c_{i}(x') \leqslant b_{i} ~\text{and}~x' \prec x \\
\end{aligned}
\end{equation*}
Similarly to the mono-objective case in \cite{gardner2014bayesian}, 
we can define the feasibility indicator function $\Delta_{i}(x) \in {0, 1}$ which is $1$ if $ c_{i}(x) \leqslant b_{i} $, and $0$ otherwise. A design point where $\Delta_{i}(x) = 1$ is termed feasible. Otherwise, it is called infeasible.

We aim to identify $\Gamma$ with the fewest possible function evaluations,
solving a sequential decision problem and constructing a strategy $\underline{X} : f \mapsto \{ X_{1}, X_{2}, X_{3}, \dots \}$
to iteratively generate the next $X_{n+1} \in \bbX$ to evaluate. 
If the evaluation $X_i$ is not very expensive then it is possible to construct a strategy that, for each sequential step, runs multiple evaluations, \ie a batch of evaluations. 
In this case it is standard practice to warm-up the strategy with some previously sampled points, using sampling techniques from the design of experiments literature~\cite{santner2013design}.

It is worth noting that, while infeasible points are never considered our best experiment, 
they are still useful to add to our set of performed experiments to improve the probabilistic model posteriors. 
Practically speaking, infeasible samples help to determine the shape and descent directions of $c(x)$, allowing the 
probabilistic model to discern which regions are more likely to be feasible without actually sampling there. 
The fact that we do not need to sample in feasible regions to find them is a property that is highly useful in cases where 
the feasible region is relatively small, and uniform sampling would have difficulty finding these regions.

As an example, in this paper, we evaluate the 
compiler optimization case for targeting FPGAs. In this case, $p = 2$, $q = 1$, $f_{1}(x) = \text{Cycles}(x)$ (number of total cycles, \ie runtime),
$f_{2}(x) = \text{Logic}(x)$ (logic utilization, \ie quantity of logic gates used) in percentage,
and $\Delta_{1}(x) \in {0, 1}$ represents whether the design point $x$ fits in the target FPGA board.


\subsection{Randomized Decision Forests}
\label{random_forests}
A decision tree is a non-parametric supervised machine learning method  widely used to formalize decision making processes across a variety of fields. 
A randomized decision tree is an analogous machine learning model, which ``learns'' how to regress (or classify) data points based on randomly selected attributes of a set of 
training examples. 
The combination of many weak regressors (binary decisions) allows approximating highly non-linear and multi-modal functions with great accuracy.
Randomized decision forests~\cite{breiman2001random,criminisi2012decision} combine many such decorrelated trees based on the randomization at the level of training data points and attributes to yield an even more effective supervised regression and classification model.

A decision tree represents a recursive binary partitioning of the input space, and uses a simple decision (a one-dimensional decision threshold) 
at each non-leaf node that aims at maximizing an ``information gain'' function. 
Prediction is performed by ``dropping'' down the test data point from the root, and letting it traverse a path decided by the node decisions, until it reaches a leaf node. 
Each leaf node has a corresponding function value (or probability distribution on function values), adjusted according to training data, which is predicted as the function value for the test input. 
During training, randomization is injected into the procedure to reduce variance and avoid overfitting.
This is achieved by training each individual tree on randomly selected subsets of the training samples (also called bagging), 
as well as by randomly selecting the deciding input variable for each tree node to decorrelate the trees. 

A regression random forests is built from a set of such decision trees where the leaf nodes output the average of the training data labels and where the output of the whole forest is the average of the predicted results over all trees. 
In our experiments, we train separate regressors to learn the mapping from our input parameter space to each output variable. 

It is believed that random forests are a good model for computer systems workloads \cite{hutter2011sequential,bodin2016integrating}. In fact, these workloads are often highly discontinuous, multi-modal, and non-linear~\cite{nardi2017algorithmic}, all characteristics that can be captured well by the space partitioning behind a decision tree. In addition, random forests naturally deal with categorical and ordinal variables which are important in computer systems optimization. Other popular models like Gaussian processes \cite{rasmussen2004gaussian} are less appealing for these type of variables. 
Additionally, a trained random forests is a ``white box'' model which is relatively simple for users to understand and to interpret (as compared to, for example, neural network models, which are more difficult to interpret).

\section{Methodology}
\label{methodology}
\subsection{Injecting Prior Knowledge to Guide the Search}
\label{probability_distribution}
Here we consider the probability densities and distributions that are useful to model computer systems workloads. In these type of workloads the following should be taken into account: 
\begin{itemize}
\item the range of values for a variable is finite. 
\item the density mass can be uniform, bell-shaped (Gaussian-like) or J-shaped (decay or exponential-like). 
\end{itemize}
For these reasons, in HyperMapper 2.0 we propose the Beta distribution as a model for the search space variables. The following three properties of the Beta distribution make it especially suitable for modeling ordinal, integer and real variables; the Beta distribution:
\begin{enumerate}
\item has a finite domain;
\item can flexibly model a wide variety of shapes including a bell-shape (symmetric or skewed), U-shape and J-shape. 
This is thanks to the parameters $\alpha$ and $\beta$ (or $a$ and $b$) of the distribution;  
\item has probability density function (PDF) given by:
\begin{equation}
	f(x|\alpha,\beta)=\frac{\Gamma(\alpha+\beta)}{\Gamma(\alpha)\Gamma(\beta)}x^{\alpha-1}(1-x)^{\beta-1} 
\end{equation}
for $x \in [0,1]$ and $\alpha, \beta > 0$, where $\Gamma$ is the Gamma function.
The mean and variance can be computed in closed form.
\end{enumerate}
Note that the Beta distribution has samples that are confined in the interval $[0,1]$. 
For ordinal and integer variables, HyperMapper 2.0 automatically rescales the samples to the range of values of the input variables and then finds the closest allowed value in the ones that define the variables. 

For categorical variables (with $K$ modalities) we use a probability distribution, \ie instead of a density, that can be easily specified as pairs of ($x_k$, $p_k$), where the set $x_k$ represents the $k$ values of the variable and $p_k$ is the probability associated to each of them with $\sum_{k=1}^{K} p_k=1$. 

\begin{figure}
\includegraphics[scale=0.7]{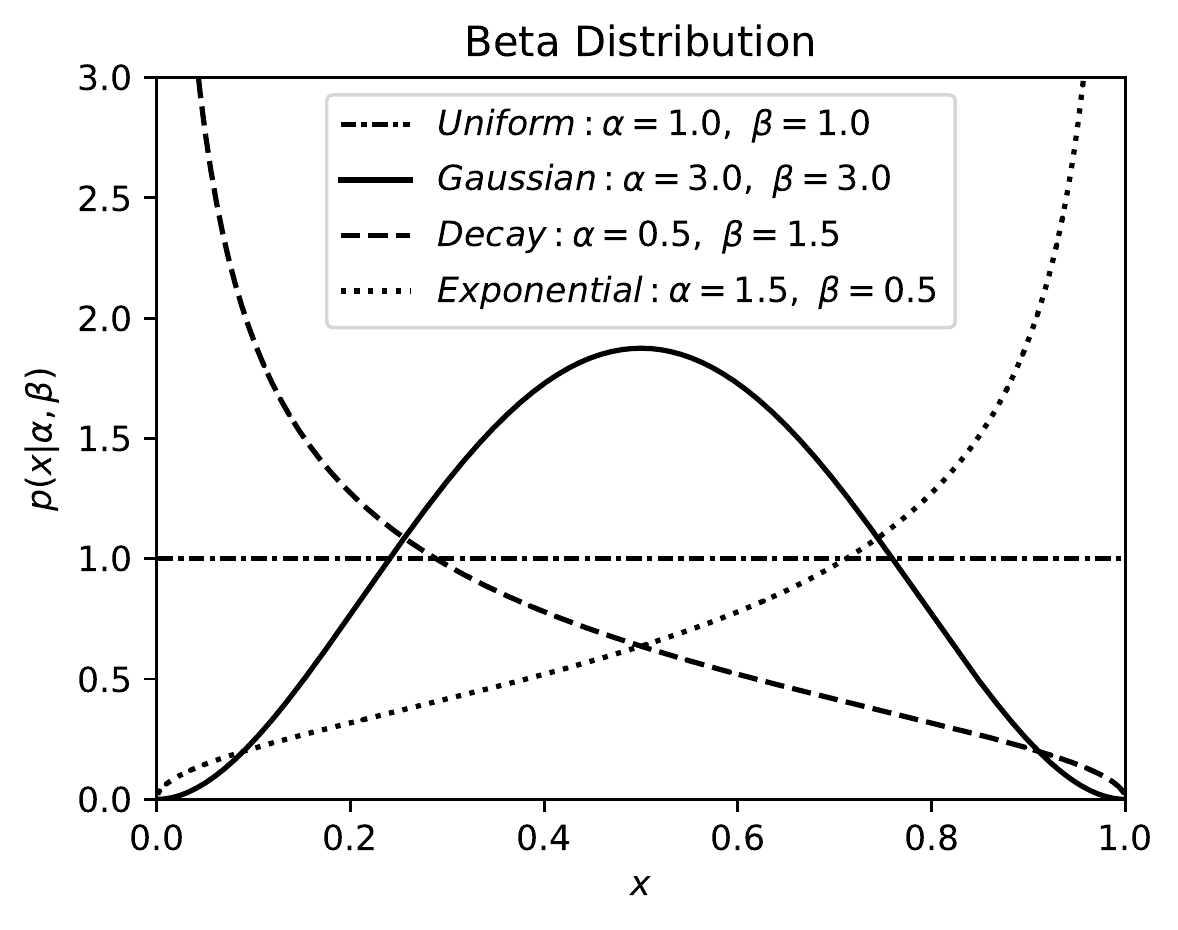}
\caption{Beta distribution shapes in HyperMapper 2.0.}
\label{Beta}
\end{figure}

In \Cref{Beta} we show Beta distributions with parameters $\alpha$ and $\beta$ selected to suit computer systems workloads.
We have selected four shapes as follows: 
\begin{enumerate}
\item Uniform ($\alpha=1,\beta=1$): used as a default if the user has no prior knowledge on the variable. 
\item Gaussian ($\alpha=3,\beta=3$): when the user thinks that it is likely that the optimum value for that variable is located in the center but still wants to sample from the whole range of values with lower probability at the borders. 
This density is reminiscent of an actual Gaussian distribution, though it is finite. 
\item Decay ($\alpha=0.5,\beta=1.5$): used when the optimum is likely located at the beginning of the range of values. 
This is similar in shape to the \textit{log-uniform} distribution as in \cite{bergstra2011algorithms,bergstra2012random}
\item Exponential ($\alpha=1.5,\beta=0.5$): used when the optimum is likely located at the end of the range of values. 
This is similar in shape to the \textit{drawn exponentially}  distribution as in \cite{bergstra2012random}
\end{enumerate}

\subsection{Sampling with Categorical and Discrete Parameters}
\label{warmup_sampling}
We first warm-up our model with simple random sampling. In the design of experiments (DoE) literature~\cite{santner2013design}, this is the most commonly used sampling technique to warm-up the search. When prior knowledge is used, samples are drawn from each variable's prior distribution, or the uniform distribution by default if no prior knowledge is provided.

\subsection{Unknown Feasibility Constraints}
\label{feasibility_constraints}
The unknown feasibility constraints algorithm in HyperMapper 2.0 is an adaptation of the constrained Bayesian optimization (cBO) method introduced in \cite{gardner2014bayesian}.
cBO is based on Gaussian Processes (GPs) to model the constraints and it uses these probabilistic models to guide the search, which is, it multiplies the acquisition function of the Bayesian optimization iteration by the constraints represented by the GPs; this leads to a new probabilistic model that is the combination of constraints and surrogate models. We refer to \cite{gardner2014bayesian} for a more detailed explanation of the cBO algorithm. 

In HyperMapper 2.0 we implement the constraints with a random forests classification model. 
The advantage of this choice is that RF is a lightweight model that is interpretable shading light on how the feasibility design space looks like. 
Experiments in Section~\ref{feasibility_classifier_performance} show the effectiveness of the random forests classifier in the context of feasibility constraints.
This model can be seen as a filter that is at the core of the search algorithm in HyperMapper 2.0, as explained in Section~\ref{active_learning}. 
The filter instructs the search algorithm on which configurations are likely to be infeasible so that the sampling budget can be used more efficiently. 

\subsection{Active Learning}
\label{active_learning}
Active learning is a paradigm in supervised machine learning which uses fewer training examples to achieve better 
prediction accuracy by iteratively training a predictor, and using the predictor in each iteration to 
choose the training examples which will increase its accuracy the most. 
Thus the optimization results are incrementally improved by interleaving exploration and exploitation steps. 
We use randomized decision forests as our base predictors created from a number of sampled points in the parameter space.

The application is evaluated on the sampled points, yielding the labels of the supervised setting given by the multiple objectives. 
Since our goal is to accurately estimate the points near the Pareto-optimal front, we use the current predictor to provide performance values 
over the parameter space and thus estimate the Pareto fronts. 
For the next iteration, 
only parameter points near the predicted Pareto front are sampled and evaluated, and subsequently used to train 
new predictors using the entire collection of training points from current and all previous iterations. This process is repeated over 
a number of iterations forming the active learning loop. Our experiments in \Cref{evaluation} indicate that this guided method of searching for 
highly informative parameter points in fact yields superior predictors as compared to a baseline that uses randomly sampled points alone. 
By iterating this process several times in the active learning loop, we are able to discover high-quality design configurations that lead to good performance outcomes.

\begin{center}
\begin{figure}[tb]
\includegraphics[scale=0.13]{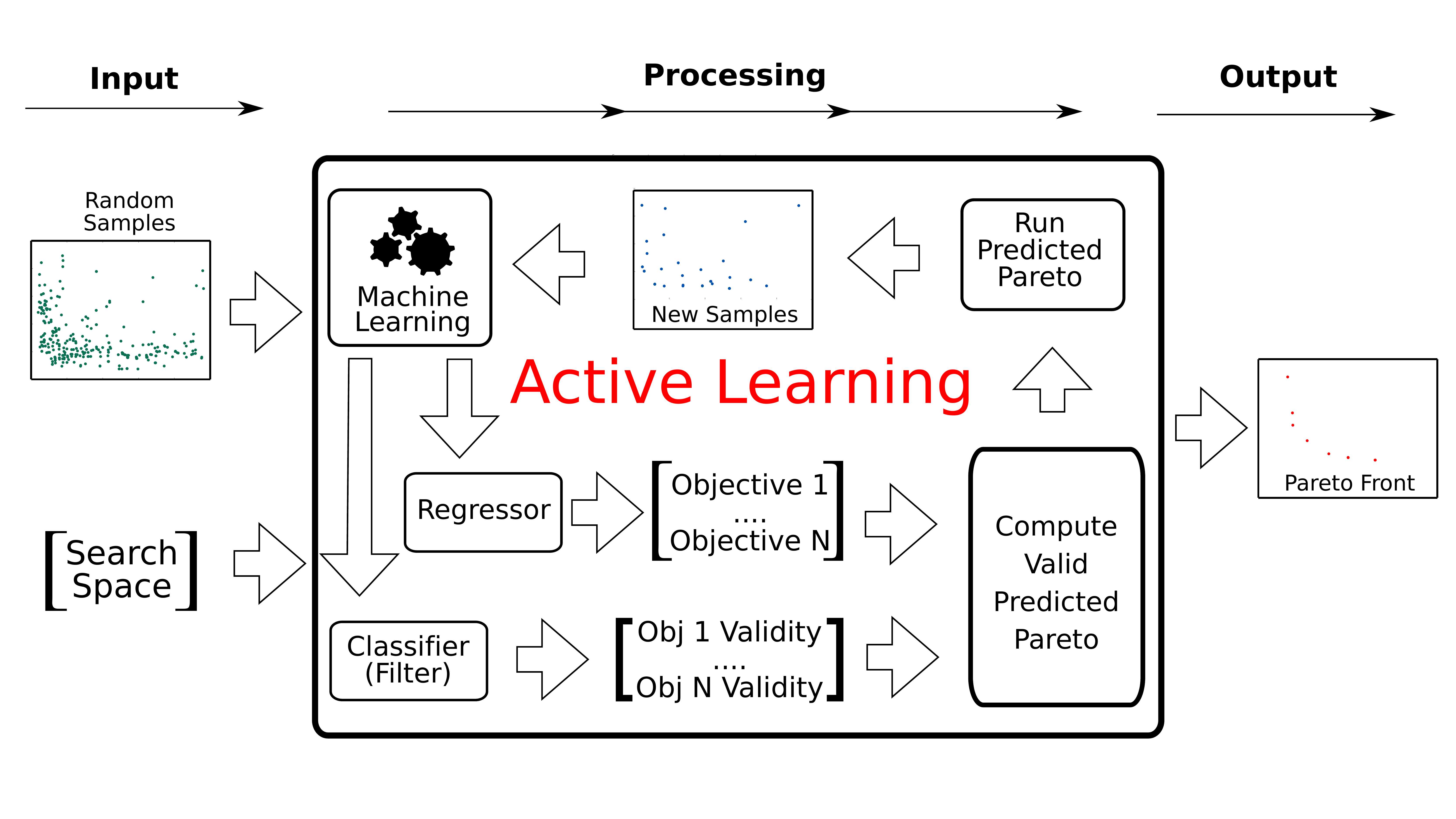}
\caption{Active learning with unknown feasibility constraints.\vspace{10pt}}
\label{figure_active_learning}
\end{figure}
\end{center}

\begin{algorithm}
 \KwData{Design space $\bbX$, warm-up sampling size $N$, $maxAL$ is the maximum samples in an active learning iteration.}
 \KwResult{Pareto front $P$.}
 Warm-up $\leftarrow$ RS; 
 
 $X_{out} \leftarrow$ Warm-up $N$ distinct configurations from $\bbX$\;
 $Y_{obj_1}, Y_{obj_2}, Y_{fea} \leftarrow Evaluate(X_{out})$\;
 $\mathcal{M}_{obj_1} \leftarrow Fit\_RF\_Regressor(X_{out}, Y_{obj_1})$\; \label{alg:fit_regressor_1}
 $\mathcal{M}_{obj_2} \leftarrow Fit\_RF\_Regressor(X_{out}, Y_{obj_2})$\; \label{alg:fit_regressor_2}
 $\mathcal{M}_{fea} \leftarrow Fit\_RF\_Classifier(X_{out}, Y_{fea})$\; \label{alg:fit_feasible_1}
 $P \leftarrow Predict\_Pareto(\mathcal{M}_{obj_1}, \mathcal{M}_{obj_2}, \mathcal{M}_{fea}, \bbX, X_{out})$\; \label{alg:predict1}
 %
 $i \leftarrow 0 $\;
 \While{$(P - X_{out} \neq \oldemptyset)$ and $(i < maxAL)$}{ \label{alg:active_learning_loop}
 	$\overline{X}_{out} \leftarrow P - X_{out}$\;
    $\overline{Y}_{obj_1}, \overline{Y}_{obj_2}, \overline{Y}_{fea} \leftarrow Evaluate(\overline{X}_{out})$\;
    $X_{out} \leftarrow X_{out} \cup \overline{X}_{out}$\;
  	$Y_{obj_1} \leftarrow Y_{obj_1} \cup \overline{Y}_{obj_1}$\;
  	$Y_{obj_2} \leftarrow Y_{obj_2} \cup \overline{Y}_{obj_2}$\;
    $Y_{fea} \leftarrow Y_{fea} \cup \overline{Y}_{fea}$\; \label{alg:fit_feasible_2}
	$\mathcal{M}_{obj_1} \leftarrow Fit\_RF\_Regressor(X_{out}, Y_{obj_1})$\; \label{alg:fit_regressor_3}
    $\mathcal{M}_{obj_2} \leftarrow Fit\_RF\_Regressor(X_{out}, Y_{obj_2})$\; \label{alg:fit_regressor_4}
 	$\mathcal{M}_{fea} \leftarrow Fit\_RF\_Classifier(X_{out}, Y_{fea})$\;
    $P \leftarrow Predict\_Pareto(\mathcal{M}_{obj_1}, \mathcal{M}_{obj_2}, \mathcal{M}_{fea}, \bbX, X_{out})$\; \label{alg:predict2}
 	$i \leftarrow i+1 $\;
 }
 \Return{$P$}\;
 \caption{Pseudo-code for HyperMapper 2.0 optimizing a two-objective ($obj_1$ and $obj_2$) application with one feasibility constraint ($fea$). $-$ denotes set difference, $\cup$ denotes set union.
}
 \label{algo:HyperMapper 2.0}
\end{algorithm}

Algorithm~\ref{algo:HyperMapper 2.0} shows the pseudo-code of the model-based search algorithm used in HyperMapper 2.0.
\Cref{figure_active_learning} shows a corresponding graphical representation of the algorithm. 
The while loop on line \ref{alg:active_learning_loop} in  Algorithm \ref{algo:HyperMapper 2.0} is the active learning loop, represented by the big loop in the preprocessing box of Figure \ref{figure_active_learning}. 
The user specifies a maximum number of active learning iterations given by the variable $maxAL$. 
The function $Fit\_RF\_Regressor()$ at lines \ref{alg:fit_regressor_1}, \ref{alg:fit_regressor_2}, \ref{alg:fit_regressor_3} and \ref{alg:fit_regressor_4} trains random forests regressors $\mathcal{M}_{obj_1}$ and $\mathcal{M}_{obj_2}$ which are the surrogate models to predict the objectives given a parameter vector. 
We train $p$ separate models, one for each objective ($p$=2 in Algorithm~\ref{algo:HyperMapper 2.0}).  
The random forests regressor is represented by the box "Regressor" in \Cref{figure_active_learning}.

The function $Fit\_RF\_Classifier()$ on lines \ref{alg:fit_feasible_1} and \ref{alg:fit_feasible_2} trains a random forests classifier $\mathcal{M}_{fea}$ to predict if a parameter vector is feasible or infeasible. 
The classifier becomes increasingly accurate during active learning. Using a classifier to predict the infeasible parameter vectors has proven to be very effective as later shown in \Cref{feasibility_classifier_performance}.  
The random forests classifier is represented by the box "Classifier (Filter)" in Figure~\ref{figure_active_learning}.
The function $Predict\_Pareto$ on lines \ref{alg:predict1} and \ref{alg:predict2} filters the parameter vectors that are predicted infeasible from $X$ before computing the Pareto, thus dramatically reducing the number of function evaluations.
This function is represented by the box "Compute Valid Predicted Pareto" in Figure \ref{figure_active_learning}.

For sake of space some details are not shown in Algorithm~\ref{algo:HyperMapper 2.0}. 
For example, the while loop on line \ref{alg:active_learning_loop} is limited to $M$ evaluations per active learning iteration. 
When the cardinality $|P - X_{out}|>M$, a maximum of $M$ samples are selected uniformly at random from the set $P - X_{out}$ for evaluation. 
In the case where $|P - X_{out}|<M$, a number of parameter vector samples $M-|P - X_{out}|$ is drawn uniformly at random without repetition. 
This ensures exploration analogous to the $\epsilon$-greedy algorithm in the reinforcement learning literature \cite{sutton1998reinforcement}. 
$\epsilon$-greedy is known to provide balance between the exploration-exploitation trade-off. 

\subsection{Pareto Wall}
\label{pareto_wall}
In Algorithm~\ref{algo:HyperMapper 2.0} lines \ref{alg:predict1} and \ref{alg:predict2}, the function $Predict\_Pareto$ eliminates the $X_{out}$ samples from $\bbX$ before computing the Pareto front.  
This means that the newly predicted Pareto front never contains previously evaluated samples and, by consequence, a new layer of Pareto front is considered at each new iteration. 
We dub this multi-layered approach the \textit{Pareto Wall} because we consider one Pareto front per active learning iteration, with the result that we are exploring several adjacent Pareto frontiers. Adjacent Pareto frontiers can be seen as a thick Pareto, \ie a Pareto Wall. 
The advantage of exploring the Pareto Wall in the active learning loop is that it minimizes the risk of using a surrogate model which is currently inaccurate. 
At each active learning step, we search previously unexplored samples which, by definition, must be predicted to be worse than the current approximated Pareto front. However, in cases where the predictor is not yet very accurate, some of these unexplored samples will often dominate the approximated Pareto, leading to a better Pareto front and an improved model.

\subsection{The HyperMapper 2.0 Framework}
\label{HyperMapper 2.0}
HyperMapper 2.0 is written in Python and makes use of widely available libraries, \eg scikit-learn and pyDOE. 
The HyperMapper 2.0 setup is via a simple json file. 
A light interface with third party software used for optimization is also necessary: templates for Python and Scala are provided. 
HyperMapper 2.0 is able to run in parallel on multi-core machines the classifiers and regressors as well as the computation of the Pareto front to accelerate the active learning iterations. 

\section{Evaluation}
\label{evaluation}

We run the evaluation on the recently proposed Spatial compiler~\cite{koeplinger2018}, which implies a full integration of HyperMapper 2.0 on the Spatial production-level compiler toolchain for designing application hardware accelerators on FPGAs. 
We compare HyperMapper 2.0 with the HyperMapper 1.0 multi-objective auto-tuner to show the effectiveness of the feasibility constraints methodology. Then we compare HyperMapper 2.0 against the real Pareto where exhaustive search is possible, i.e. a total of three benchmarks. This is to give an insight on how the optimizer works in a controlled environment, i.e. when the Pareto front in known and the benchmark is small. Finally a comparison with the previous approach in Spatial, which is a mix of expert programmer pruning and random sampling, is given. The blocking factor to run more comparisons with other auto-tuners is that there is no available framework that has both RIOC variables and multi-objective features as shown in Table~\ref{table_taxonomy_software}. 
As an example, the popular OpenTuner supports multiple objectives that are scalarized in one objective or allows optimization of one objective while thresholding a second objective. This means that OpenTuner is inherently single-objective\footnote{Refer to this OpenTuner code for example: https://tinyurl.com/ybxcokyn.} making the comparison with HyperMapper 2.0 not legitimate.

\subsection{The Spatial Programming Language}
\label{Spatial}

\begin{figure}
\includegraphics[width=\columnwidth]{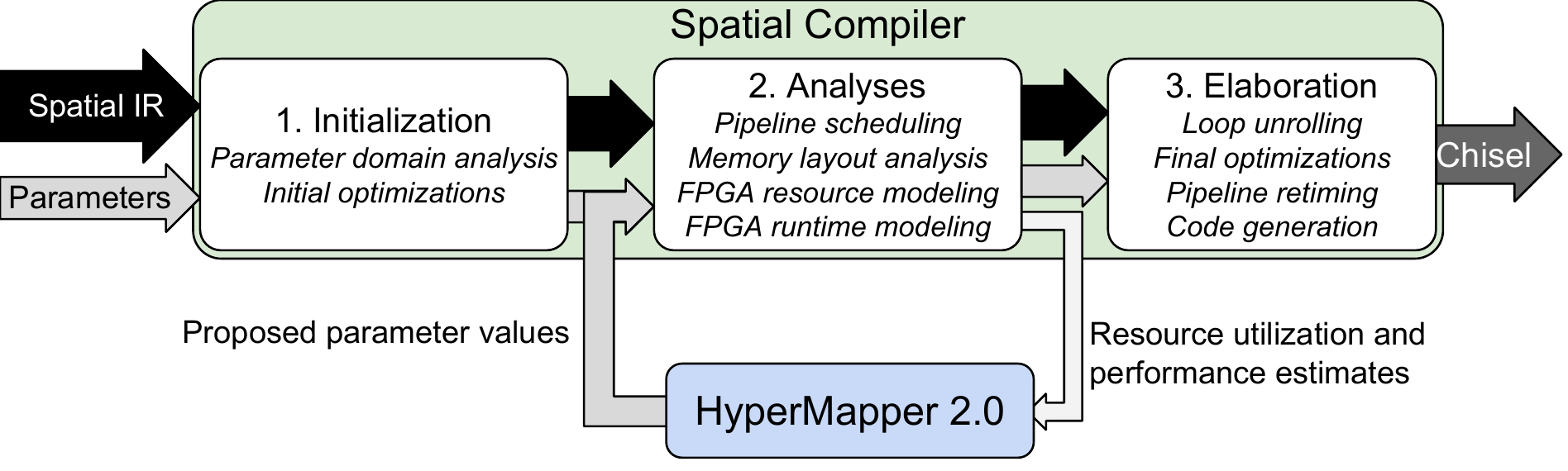}
\caption{An overview of the phases of the compiler for the Spatial application accelerator design language. HyperMapper 2.0 interfaces at the beginning and ending of phase 2 to drive accelerator design space exploration and the selection of design parameter values.}
\label{fig:spatial}
\end{figure}

Spatial~\cite{koeplinger2018} is a domain-specific language (DSL) and corresponding compiler for the design of application accelerators on reconfigurable architectures. The Spatial frontend is tailored to present programmers with a high level of abstraction for hardware design. Control in Spatial is expressed as nested, parallelizable loops, while data structures are allocated based on their placement in the target hardware's memory hierarchy. The language also includes support for design parameters to express values which do not change the behavior of the application and which can be changed by the compiler. These parameters can be used to express loop tile sizes, memory sizes, loop unrolling factors, and the like. 

As shown in \Cref{fig:spatial}, the Spatial compiler lowers user programs into synthesizable Chisel~\cite{chisel} designs in three phases. In the first phase, it performs basic hardware optimizations and estimates a possible domain for each design parameter in the program. In the second phase, the compiler computes loop pipeline schedules and on-chip memory layouts for some given value for each parameter. It then estimates the amount of hardware resources and the runtime of the application. When targeting an FPGA, the compiler uses a device-specific model to estimate the amount of compute logic (LUTs), dedicated multipliers (DSPs), and on-chip scratchpads (BRAMs) required to instantiate the design. Runtime estimates are performed using similar device-specific models with average and worst case estimates computed for runtime-dependent values. Runtime is typically reported in clock cycles. 

In the final phase of compilation, the Spatial compiler unrolls parallelized loops, retimes pipelines via register insertion, and performs on-chip memory layout and compute optimizations based on the analyses performed in the previous phase. Finally, the last pass generates a Chisel design which can be synthesized and run on the target FPGA.

\begin{table}[thb]
    \begin{center}
    \begin{small}
      \begin{tabular}{ | l | c | l |}
	\hline
    \textbf{Benchmark} & \textbf{Variables} & \textbf{Space Size} \\ \hline\hline
	BlackScholes & 4 & $7.68 \times 10^4$ \\ \hline
    K-Means & 6 & $1.04 \times 10^6$ \\ \hline
    OuterProduct & 5 & $1.66 \times 10^7$ \\ \hline
    DotProduct & 5 & $1.18 \times 10^8$ \\ \hline
    GEMM & 7 & $2.62 \times 10^8$ \\ \hline
    TPC-H Q6 & 5 & $3.54 \times 10^9$ \\ \hline
    GDA & 9 & $2.40 \times 10^{11}$ \\ \hline
\end{tabular}
    \end{small}
    \end{center}
  \caption{Spatial benchmarks and design space size.}
  \label{table_spatial_benchmarks}
\end{table}


\subsection{HyperMapper 2.0 in the Spatial Compiler}
\label{Spatial-ParetoDaemon_compiler}
The collection of design parameters in a Spatial program, together with their respective domains, yields a hardware design space. The second phase of the compiler gives a way to estimate two cost metrics - performance and FPGA resource utilization - for a given design in this space. Existing work on Spatial has evaluated two methods for design space exploration. The first method heuristically prunes the design space and then performs randomized search with a fixed number of samples. The heuristics, first established by Spatial's predecessor~\cite{dhdl}, help to eliminate obviously bad points within the design space prior to random search; 
the pruning is provided by expert FPGA developers. This is, in essence, a one-time hint to guide search.  
The second method evaluated the feasibility of using HyperMapper 1.0~\cite{bodin2016integrating} to drive exploration, concluding that the tool was promising but still required future development. 
In some cases, it performed poorly without a feasibility classifier as the search often focused on infeasible regions~\cite{koeplinger2018}. 

Spatial's compiler includes hooks at the beginning and end of its second phase to interface with external tools for design space exploration. As shown in \Cref{fig:spatial}, the compiler can query at the beginning of this phase for parameter values to evaluate. Similarly, the end of the second phase has hooks to output performance and resource estimates. HyperMapper 2.0 interfaces with these hooks to receive cost estimates, build a surrogate model, and drive search of the space. 

In this work, we evaluate design space exploration when Spatial is targeting an Altera Stratix V FPGA with 48 GB of dedicated DRAM and a peak memory bandwidth of 76.8~GB/sec (an identical approach could be used for any FPGA target). We list the seven benchmarks we evaluate with HyperMapper 2.0 in \Cref{table_spatial_benchmarks}. These seven benchmarks are a representative subset of those previously used to evaluate the Spatial compiler~\cite{koeplinger2018}.  

\subsection{Feasibility Classifier Effectiveness}
\label{feasibility_classifier_performance}
We address the question of the effectiveness of the feasibility classifier in the Spatial use case. 
Of all the hyperparameters defined for binary random forests~\cite{scikit-learn}, the parameters that usually have the most impact on the performance of the random forests classifier are: $n\_estimators$, $max\_depth$, $max\_features$ and $class\_weight$. 
The reader car refer to the scikit-learn random forests classifier documentation for more details on these model hyperparameters. 
We run an exhaustive search to fine-tune the binary random forests classifier hyperparameters and test its performance. 
The range of values we considered for these parameters is shown in \Cref{tab_exhaustive_search_space_binary_classifier}.
This defines a comprehensive space of 81 possible choices, small enough that it can be explored using exhaustive search. We dub these choices  of parameter vectors as $config\_1$ to $config\_81$ on the $x$ axis. 

\begin{table}[thb]
    \begin{center}
    \begin{footnotesize}
      \begin{tabular}{ | l | l | }
	\hline
    Name & Range of Values \\ \hline
	\hline
	$n\_estimators$ &  [10, 100, 1000] \\ \hline
    $max\_depth$ & [None, 4, 8]  \\ \hline
    $max\_features$ & ['auto', 0.5, 0.75]  \\ \hline
    $class\_weight$ & [\{T: 0.50, F: 0.50\}, \{T: 0.75, F: 0.25\}, \{T: 0.9, F: 0.1\}] \\ \hline
      \end{tabular}
    \end{footnotesize}
    \end{center}
  \caption{Random forests classifier hyperparameter tuning search space.  
  }
  \label{tab_exhaustive_search_space_binary_classifier}
\end{table}
We perform a 5-fold cross-validation using the data collected by HyperMapper 2.0 
as training data and report validation recall averaged over the 5 folds. 
The goal of this optimization procedure is for the binary classification to maximize recall.
We want to maximize recall, \ie $\frac{true\ positives}{true\ positives+false\ negatives}$, because it is important to not throw away feasible points that are misclassified as being infeasible and that can potentially be good fits. 
Precision, \ie $\frac{true\ positives}{true\ positives + false\ positives}$, is less important as there is smaller cost associated with classifying an infeasible parameter vector as feasible. 
In this case the only downside is that some samples will be wasted because we are evaluating samples that are infeasible, which is not a major drawback.


\begin{figure}
\centering

\subfloat[Recall at 0 active learning iterations.]{
\begin{minipage}{\linewidth}
\begin{tikzpicture}
  \node (img)  {\includegraphics[clip, trim=5.5cm 14.3cm 2cm 0.0cm, width=0.91\linewidth]{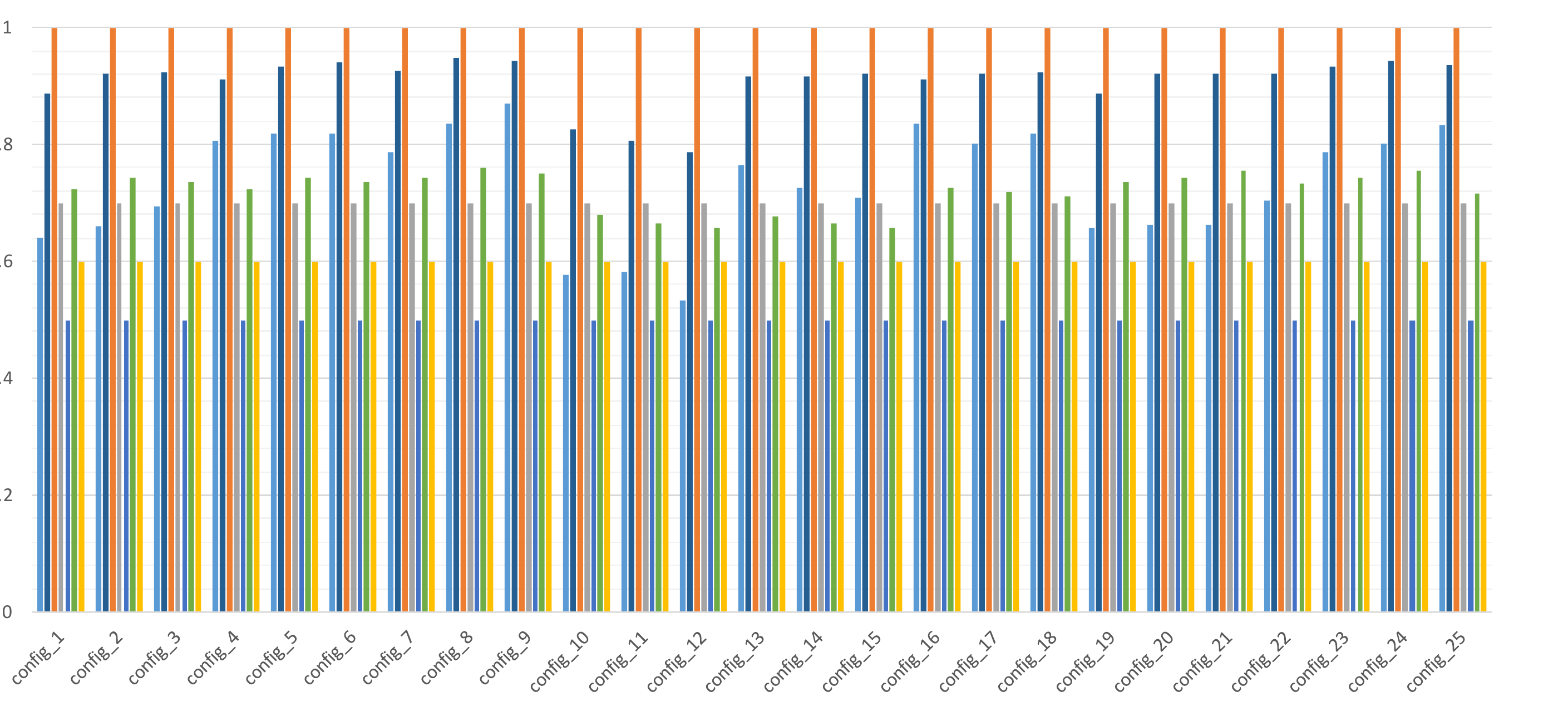}
};
  \node[left=of img, node distance=0cm, rotate=90, anchor=center,yshift=-0.7cm,font=\color{black}] {Recall};
\end{tikzpicture}
\centering\begin{tabular}{ | l | l | l | }
\hline
\textbf{Max mean} & \textbf{Max median} & \textbf{Max min} \\ \hline
\hline
0.784 & 0.826 & 0.600 \\ \hline
\end{tabular}

\end{minipage}
\label{fig:feasibility_recall_AL0}
}

\subfloat[Recall at 50 active learning iterations.]{
\begin{minipage}{\linewidth}
\begin{tikzpicture}
  \node (img)  {\includegraphics[clip, trim=5.5cm 1.5cm 2cm 13cm, width=0.91\linewidth]{RF_classifier_CV_performance.pdf}
};
  \node[left=of img, node distance=0cm, rotate=90, anchor=center,yshift=-0.7cm,font=\color{black}] {Recall};
\end{tikzpicture}
\centering
\begin{tabular}{ | l | l | l | }
\hline
\textbf{Max mean} & \textbf{Max median} & \textbf{Max min} \\ \hline
\hline
0.967 & 0.984 & 0.886 \\ \hline
\end{tabular}
\end{minipage}%
}
\vspace{5pt}
\centering\includegraphics[clip, trim=5.5cm 0.1cm 2.5cm 26cm, width=\linewidth]{RF_classifier_CV_performance.pdf}

\caption{RF feasibility  classifier 5-fold cross-validation recall over all benchmarks. The first 25 hyperparameter configurations of the classifier are shown. ``Max mean'', ``Max median'', and ``Max min'' are the maximum across the mean, median, and minimum recall scores for all 7 benchmarks, respectively.}
\label{figure_feasibility_classification}
\end{figure}

\Cref{figure_feasibility_classification} reports the recall of the random forests classifier across the 7 benchmarks and hyperparameter configurations. 
For sake of space, we only report the first 25 configurations, but the trend persists across all configurations.
\Cref{figure_feasibility_classification} (top) shows the recall just after the warm-up sampling and before the first active learning iteration (Algorithm~\ref{algo:HyperMapper 2.0} line~\ref{alg:fit_feasible_1}). Figure~\ref{figure_feasibility_classification} (bottom) shows the recall after 50 active learning iterations (Algorithm~\ref{algo:HyperMapper 2.0} line~\ref{alg:fit_feasible_2} after 50 iterations of the while loop, where each iteration is evaluating 100 samples). 
The recall goes up during the active learning loop implying that the feasibility constraint is being predicted more accurately over time. 
The tables in \Cref{figure_feasibility_classification} show this  general performance trend with the max mean improving from 0.784 to 0.967.  


In \Cref{figure_feasibility_classification} (top) the recall is low prior to the start of active learning. 
The configuration that scores best (the maximum score of the minimum scores across the different configurations) has a minimum score of 0.6 on the 7 benchmarks. 
The configuration is: 
\{'class\_weight':\{T:0.75,F:0.25\}, 'max\_depth':8, 'max\_features':'auto', 'n\_estimators':10\}. 
The recall of this configuration ranges from a minimum of 0.6 for TPC-H Q6 to a maximum of 1.0 on BlackScholes with mean and standard deviation of 0.735 and 0.15 respectively. 

In \Cref{figure_feasibility_classification} (bottom)  the recall is high after 50 iterations of active learning. 
There are two configurations that score best, with a minimum score of 0.886 on the 7 benchmarks. 
The configurations are: 
\{'class\_weight':\{T:0.75,F:0.25\}, 'max\_depth':None, 'max\_features':'0.75', 'n\_estimators':10\} and 
\{'class\_weight':\{T:0.9,F:0.1\}, 'max\_depth':None, 'max\_features':'0.75', 'n\_estimators':10\}. 
In general, most of the configurations are very close in terms of recall and the default random forests configuration scores high, perhaps suggesting that the random forests for these kind of workloads does not need a major tuning effort. 
The statistics of these configurations range from a minimum of 0.886 for DotProduct to a maximum of 1.0 on BlackScholes with mean and standard deviation of 0.964 and 0.04 respectively. 
 
\begin{figure}
\includegraphics[width=\columnwidth]{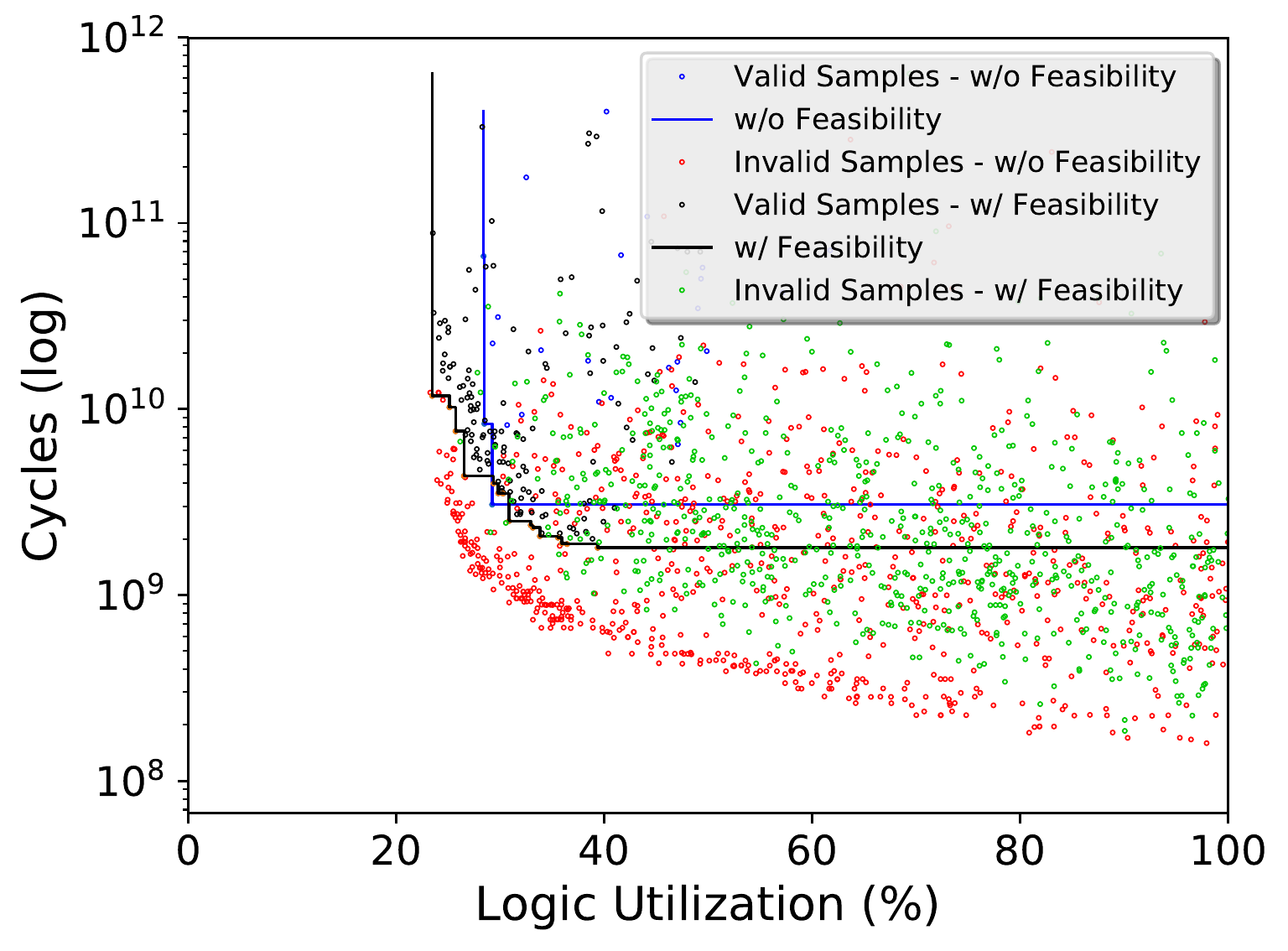}
\caption{Effect of the binary constraint classifier on the GDA benchmark. 
HyperMapper 2.0, with his feasibility classifier feature, is shown as black (feasible) and green (infeasible). HyperMapper 1.0 is shown as blue (feasible) and red (infeasible). The final approximated Pareto fronts are shown as black (our approach) and blue (HyperMapper 1.0) curves. 
}
\label{figure_classifier_pareto}
\end{figure}

\Cref{figure_classifier_pareto} compares the predicted Pareto fronts of GDA, the benchmark with the largest design space, using HyperMapper 1.0 and HyperMapper 2.0. HyperMapper 1.0 does not exploit the feasibility constraints feature introduced by HyperMapper 2.0 as shown in Table~\ref{table_taxonomy_software}.
In both cases, we use random sampling to warm-up the optimization with 1,000 samples followed by 5 iterations of active learning.  
The red dots representing the invalid points for the case without feasibility constraints (HyperMapper 1.0) are spread farther from the corresponding Pareto frontier while the green dots for the case with constraints (HyperMapper 2.0) are close to the respective frontier. 
This happens because the non-constrained search focuses on seemingly promising but unrealistic points. 
HyperMapper 2.0 with its constrained search focuses in a region that is more conservative but feasible. 
The effect of the feasibility constraint is apparent in its improved Pareto front, which almost entirely dominates the approximated Pareto front resulting from unconstrained search. 
For the sake of brevity, we only show experiments on the biggest design space considered in this evaluation section, i.e., the GDA benchmark, however the results are confirmed in the rest of the benchmarks.  

\subsection{Optimum vs. Approximated Pareto}
\label{optimum_vs_approximated_pareto_front}
We next take the smallest benchmarks, BlackScholes, DotProduct and OuterProduct, and run exhaustive search to compare the approximated Pareto front computed by HyperMapper 2.0 with the true optimal one. This can be achieved only for such small benchmarks as exhaustive search is feasible. However, even on these small spaces, exhaustive search requires 6 to 12 hours when parallelized across 16 CPU cores. In our framework, we use random sampling to warm-up the search with \num[group-separator={,}]{1000} random samples followed by 5 active learning iterations of about 500 samples total. 

\begin{center}
\begin{figure*}[thb]
\minipage{0.33\textwidth}
\centering
\includegraphics[width=\linewidth]{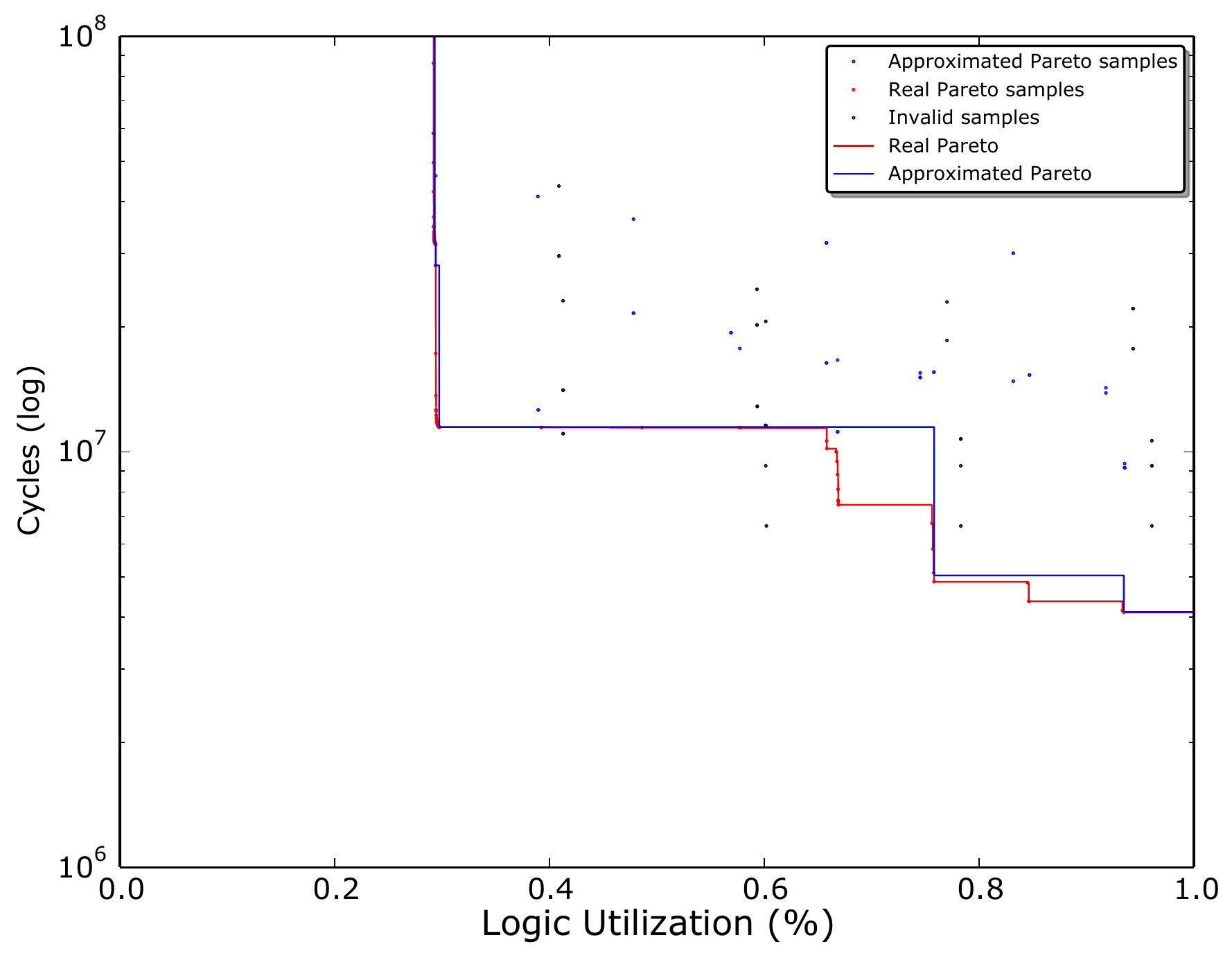}\\
BlackScholes
\endminipage\hfill
\minipage{0.33\textwidth}
\centering
\includegraphics[width=\linewidth]{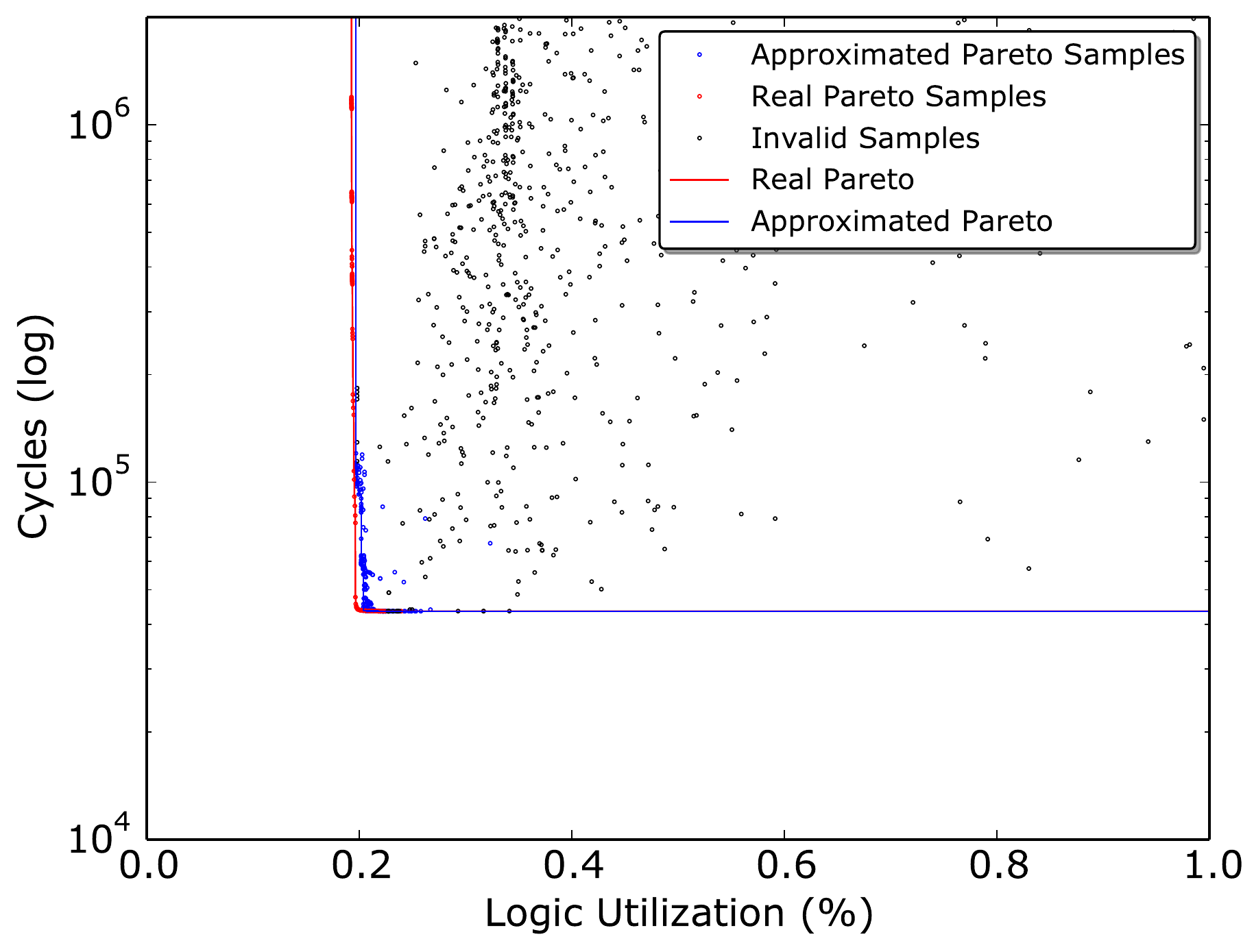}\\
OuterProduct
\endminipage\hfill
\minipage{0.33\textwidth}
\centering
\includegraphics[width=\linewidth]{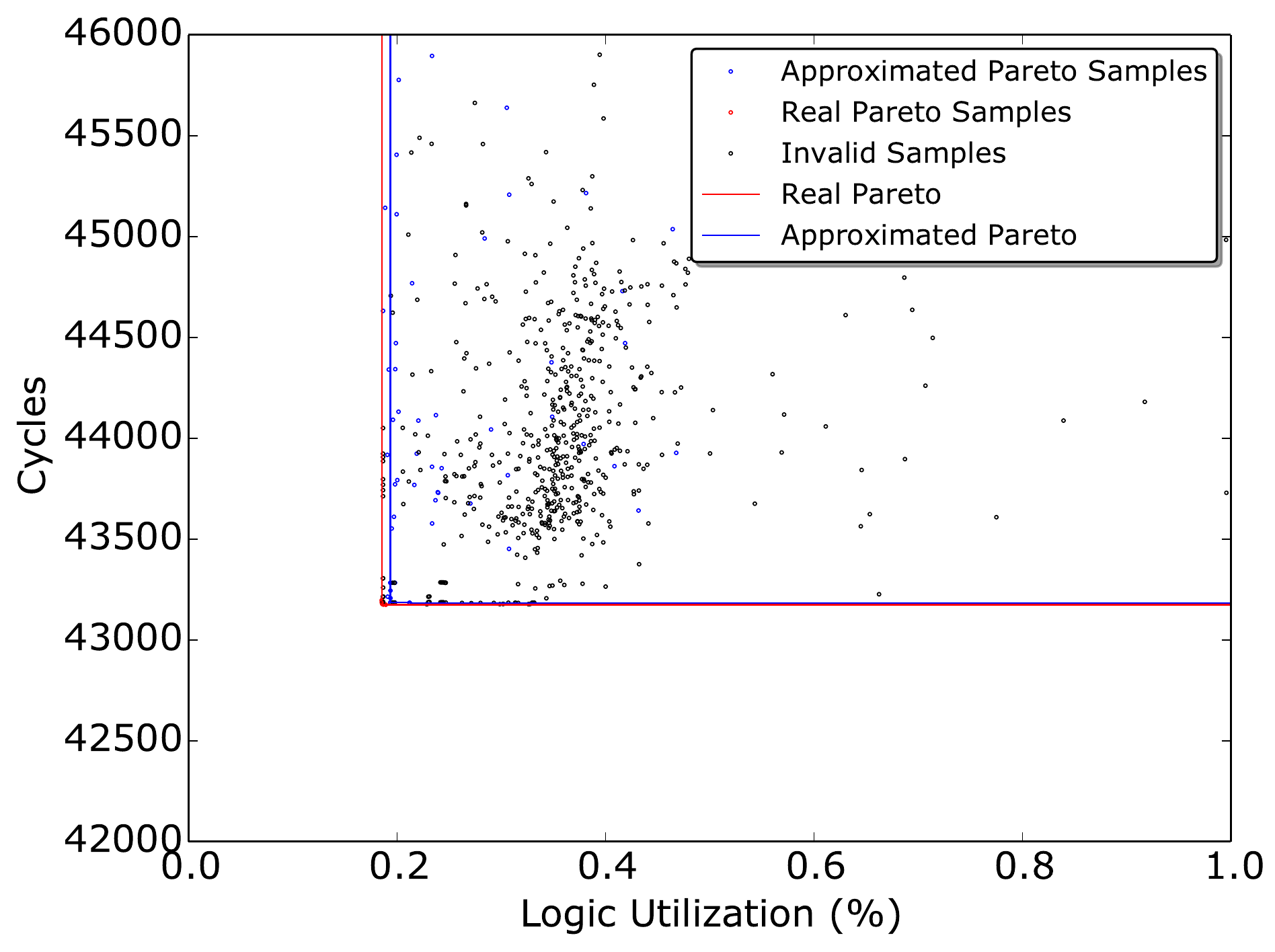}\\
DotProduct
\endminipage
\caption{Optimum versus approximated Pareto front for the BlackScholes (left, y-axis in log scale), OuterProduct (center, y-axis in log scale) and DotProduct (right) benchmarks. The x-axis is compute logic, reported as a percentage of the total LUT capacity of the Stratix V FPGA. The y-axis is the total cycles taken to run the benchmark. The approximated Pareto front is computed by HyperMapper 2.0 and the real Pareto is computed by exhaustive search. The invalid (or infeasible) samples are samples that would not be possible to synthesize on the FPGA given the hardware constraints.}
\label{optimum_pareto} 
\end{figure*}
\end{center}

Comparisons are synthesized in \Cref{optimum_pareto}. 
The optimal Pareto front is very close to the approximated one provided by HyperMapper 2.0, showing our software's ability to recover the optimal Pareto front. 
About \num[group-separator={,}]{1500} total samples are required to recover the Pareto optimum, about the same number of samples for BlackScholes and 66 times fewer for OuterProduct and DotProduct compared to the prior Spatial design space exploration approach using pruning and random sampling. 

\subsection{Hypervolume Indicator}
\label{hvi}
We next show the hypervolume indicator (HVI) \cite{feliot2017bayesian} 
function for the whole set of the Spatial benchmarks as a function of the initial number of warm-up samples (for sake of space we omit the smallest benchmark, BlackScholes). 
For every benchmark, we show 5 repetitions of the experiments and report variability via a line plot with 80\% confidence interval. 
The HVI metric gives the area between the estimated Pareto frontier and the space’s true Pareto front. 
This metric is the most common to compare multi-objective algorithm performance. 
Since the true Pareto front is not always known, we use the accumulation of all experiments run on a given benchmark to compute our best approximation of the true Pareto front and use this as a true Pareto. This includes all repetitions across all approaches, \eg baseline and HyperMapper 2.0.
In addition, since logic utilization and cycles have different value ranges by several order of magnitude, we normalize the data by dividing by the standard deviation before computing the HVI.  
This has the effect of giving the same importance to the two objectives and not skewing the results towards the objective with higher raw values. 
We set the same number of samples for all the experiments to \num[group-separator={,}]{100000}
(the default value in the prior work baseline).
Based on advice by expert hardware developers, we modify the Spatial compiler to automatically generate the prior knowledge discussed in \Cref{probability_distribution} based on design parameter types. For example, on-chip tile sizes have a ``decay'' prior because increasing memory size initially helps to improve DRAM bandwidth utilization but has diminishing returns after a certain point.
This prior information is passed to HyperMapper 2.0 and is used to magnify random sampling. The baseline has no support for prior knowledge. 

\begin{center}
\begin{figure*}[thb]
\minipage{0.33\textwidth}
\centering
\includegraphics[width=\linewidth]{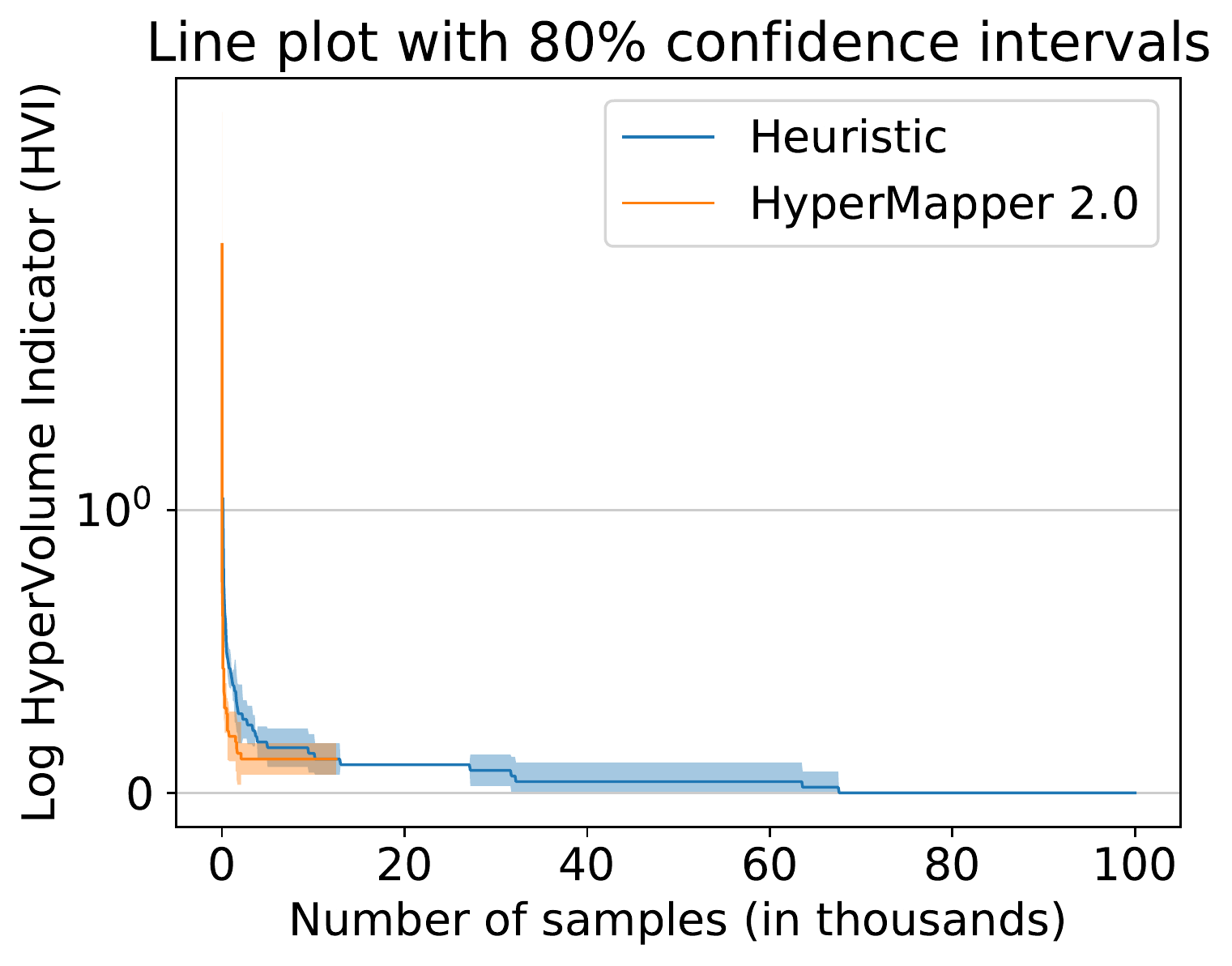}\\
GEMM
\endminipage\hfill
\minipage{0.33\textwidth}
\centering
\includegraphics[width=\linewidth]{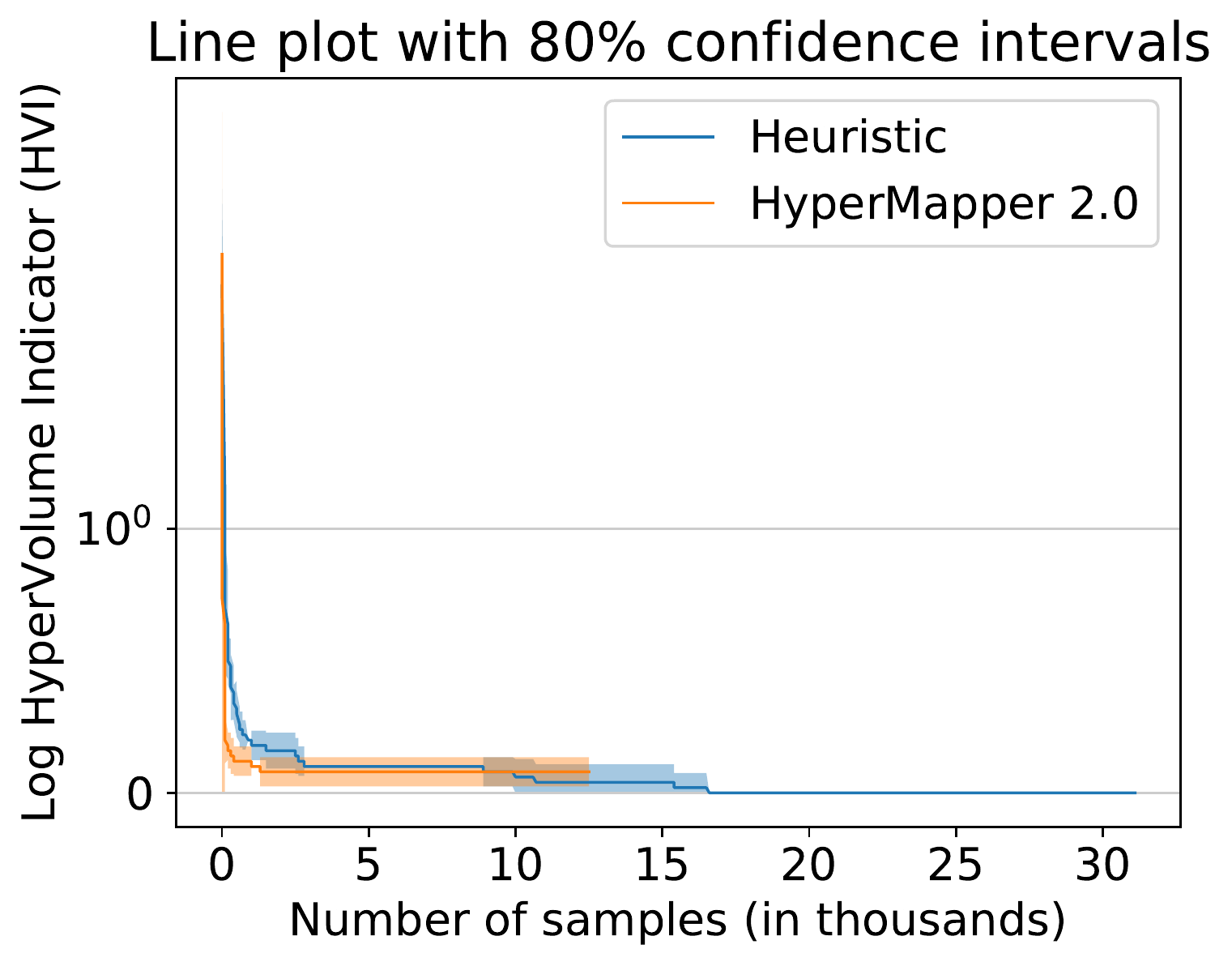}\\
OuterProduct
\endminipage\hfill
\minipage{0.33\textwidth}
\centering
\includegraphics[width=\linewidth]{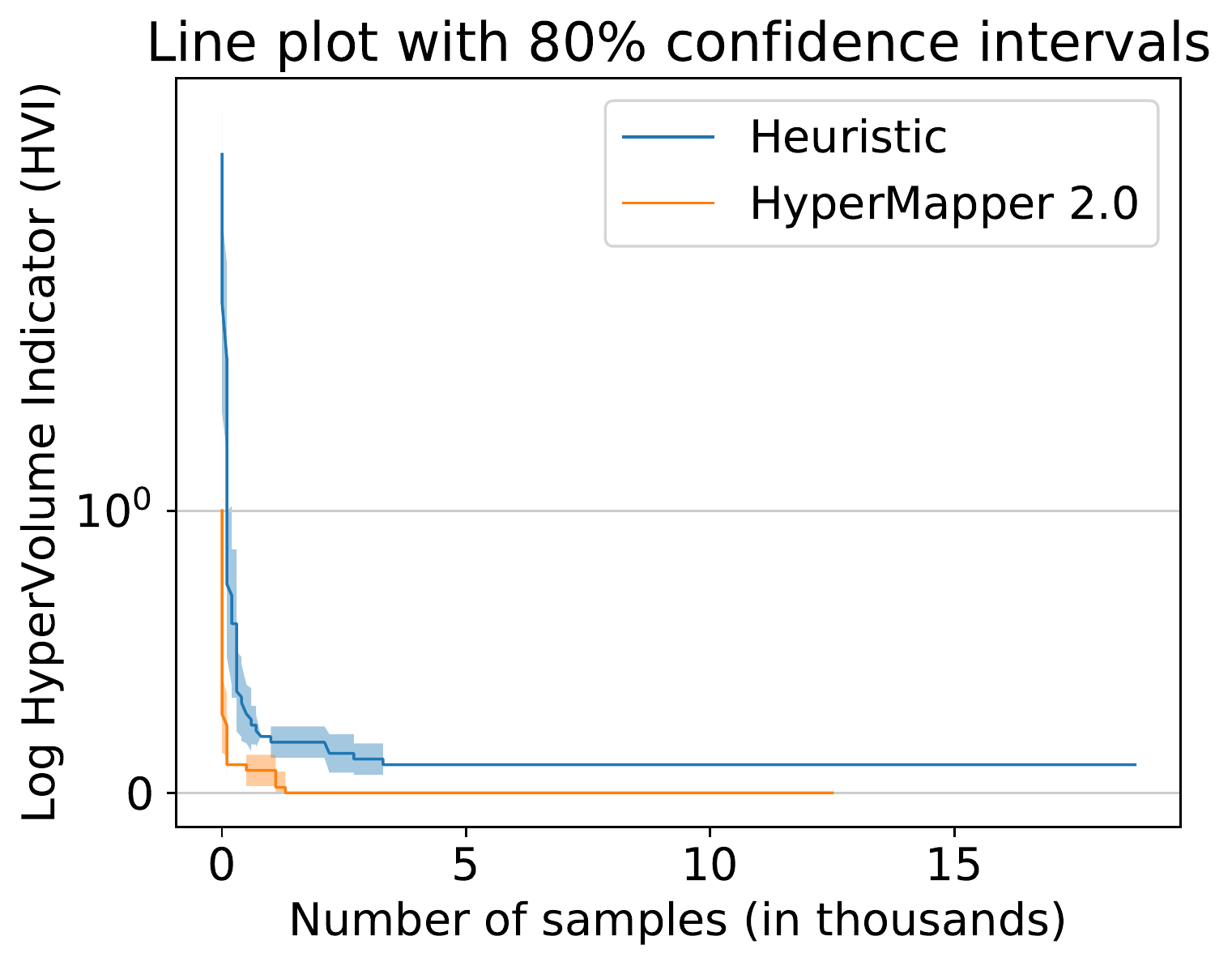}\\
K-Means
\endminipage\hfill
\minipage{0.33\textwidth}
\centering
\includegraphics[width=\linewidth]{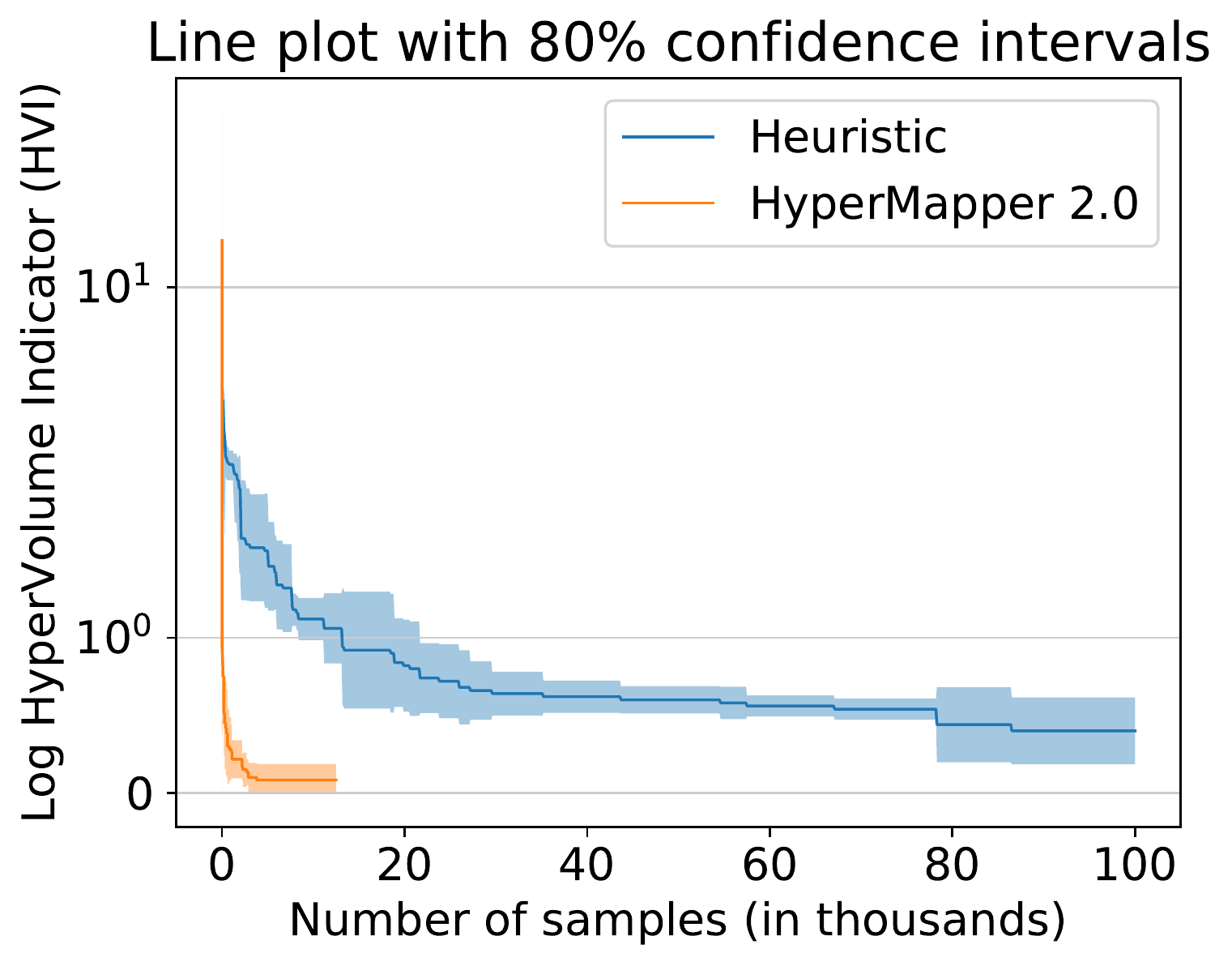}\\
GDA
\endminipage\hfill
\minipage{0.33\textwidth}
\centering
\includegraphics[width=\linewidth]{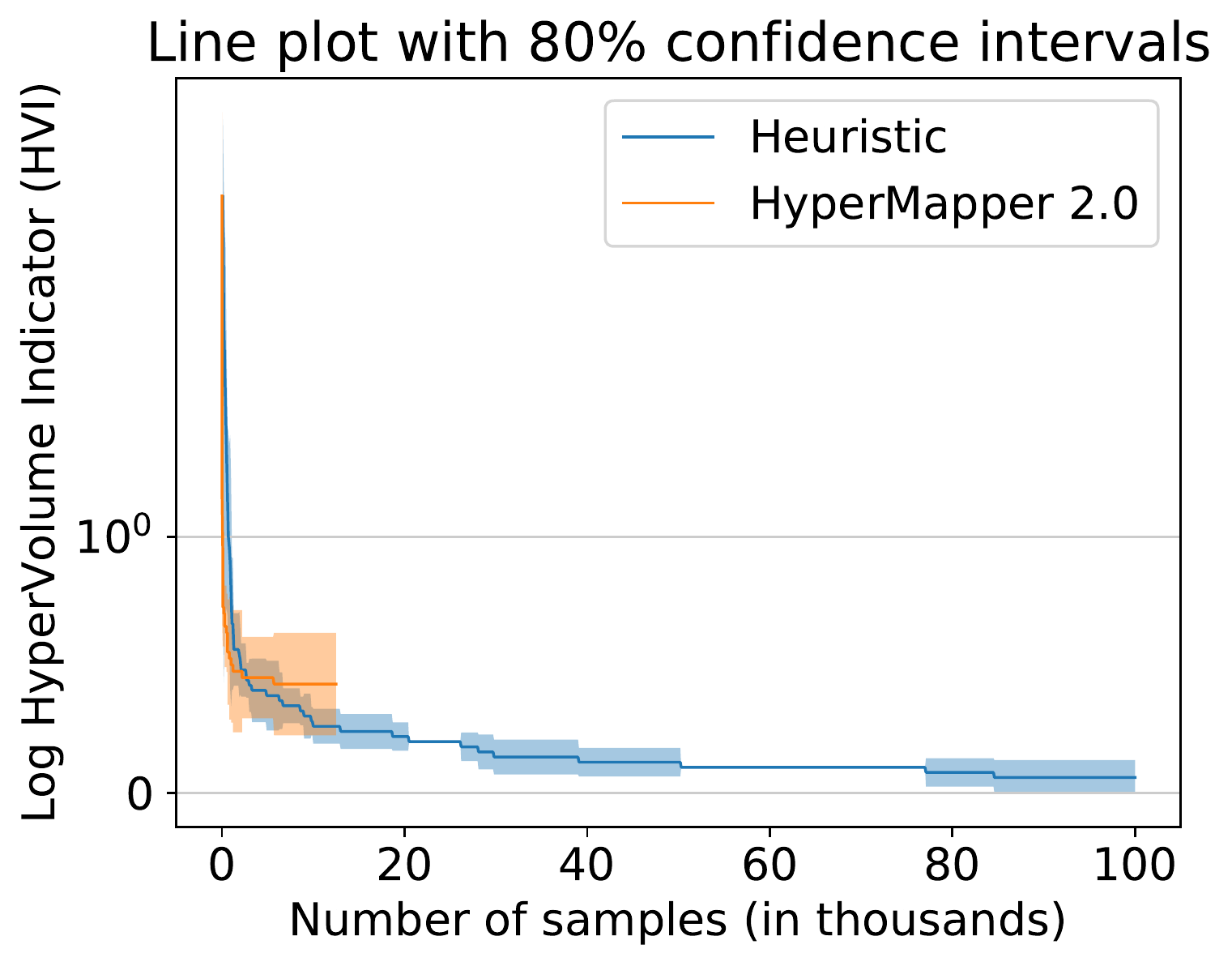}\\
T-PCH Q6
\endminipage\hfill
\minipage{0.33\textwidth}
\centering
\includegraphics[width=\linewidth]{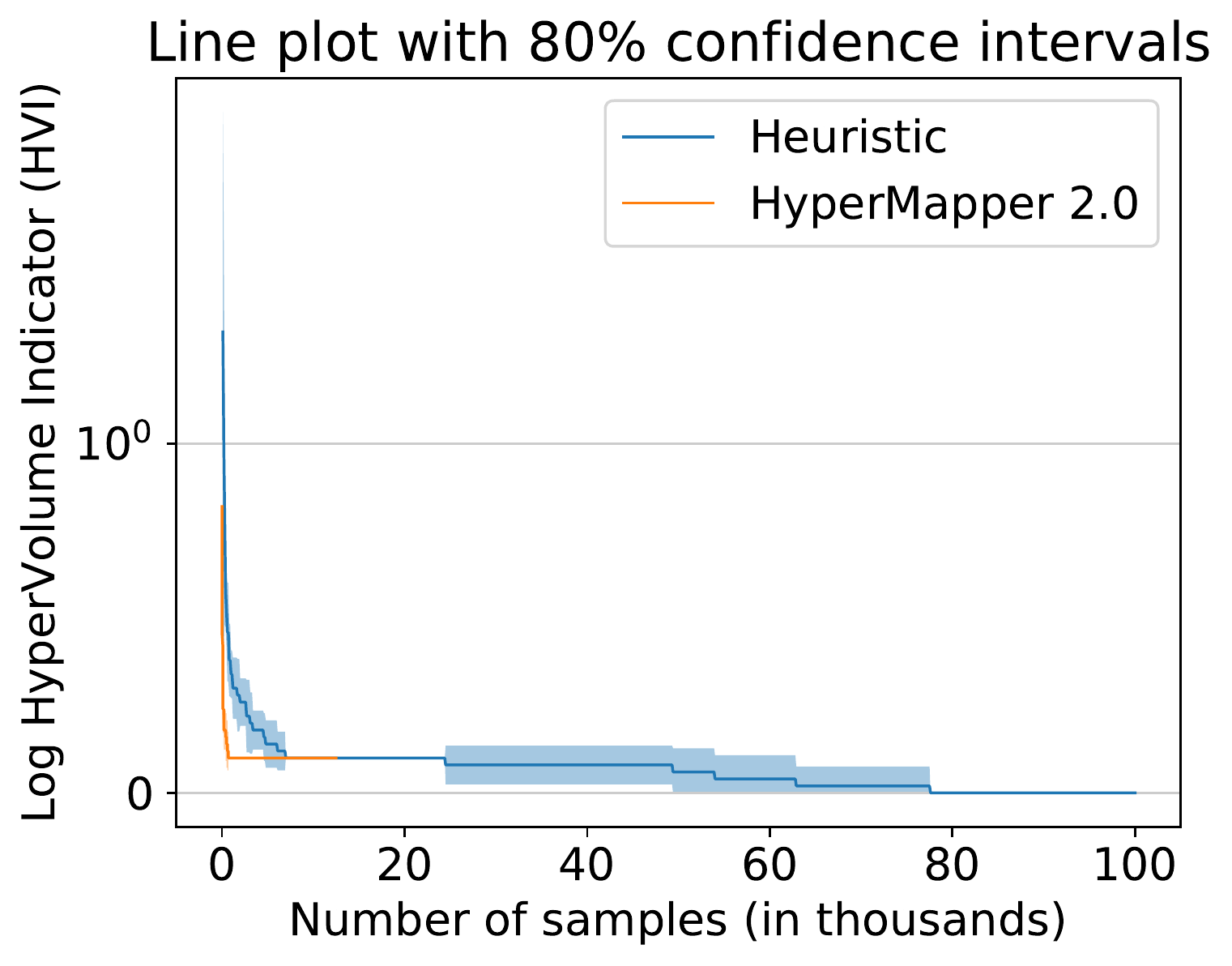}\\
DotProduct
\endminipage
\caption{Performance of HyperMapper 2.0 versus the Spatial programming language  baseline. }
\label{hvi_rs_al} 
\end{figure*}
\end{center}

\Cref{hvi_rs_al} shows the two different approaches: 
HyperMapper 2.0 using a warm-up sampling phase with the use of the prior and then an active learning phase;  
Spatial's previous design space exploration approach (the baseline). 
The y-axis reports the HVI metric and the x-axis the number of samples in thousands. 

\Cref{table_performance} quantitatively summarizes the results.
We observe the general trend that HyperMapper 2.0 needs far fewer samples to achieve competitive performance compared to the baseline. Additionally, our framework's variance is generally small, as shown in \ref{hvi_rs_al}.
The total number of samples used by HyperMapper 2.0 is 12,500 on all experiments while the number of samples performed by the baseline varies as a function of the pruning strategy. 
The number of samples for GEMM, T-PCH Q6, GDA, and DotProduct is 100,000, which leads to an efficiency improvement of $8\times$,
while OuterProduct and K-Means are 31,068 and 18,720, which leads to an improvement of $2.49\times$ and $1.5\times$, respectively.


\begin{table}
    \begin{center}
    \begin{small}
      \begin{tabular}{ | l | l | l |}
	\hline
    \textbf{Benchmark} & \textbf{HyperMapper 2.0} & \textbf{Spatial Baseline} \\ \hline\hline
	BlackScholes & $ 0.1 \pm 0$ & $0 \pm 0$ \\ \hline
    K-Means & $0 \pm 0 $ & $ 0.1 \pm 0.05$ \\ \hline
    OuterProduct & $ 0.08 \pm 0.06$ & $ 0 \pm 0$\\ \hline
    DotProduct & $ 0.1 \pm 0$ &  $0 \pm 0$\\ \hline
    GEMM & $ 0.12 \pm 0.06$& $ 0 \pm 0$\\ \hline
    TPC-H Q6 & $0.43 \pm 0.2$ &$ 0.06 \pm 0.06 $\\ \hline
    GDA & $0.08 \pm 0.11$ & $0.4 \pm 0.22$ \\ \hline
\end{tabular}
    \end{small}
    \end{center}
  \caption{Performance of HyperMapper 2.0 in terms of mean $\pm$~80\% confidence interval at the end of the optimization process. 
  Note that our approach terminates using a much lower number of samples.}
  \label{table_performance}
\end{table}



As a result, the autotuner is robust to randomness and only a reasonable number of random samples are needed for the warm-up and active learning phases. 

\subsection{Understandability of the Results}
\label{understandability}

\begin{table}
\centering
\begin{tabular}{| l | r | c | c |}
\hline
						& 					   	   & \multicolumn{2}{c|}{\bf{Objective}} \\
\bf{Benchmark}     & \bf{Parameter} 	   	   & \bf{Logic Util.}  & \bf{Cycles}  \\ \hline
\multirow{4}{*}{BlackScholes}   & \emph{Tile Size}     & 0.003			   & \bf{0.569}	  \\ 
							    & \emph{OP}			   & 0.261			   & 0.072 		  \\
							    & \emph{IP}			   & \bf{0.735}		   & 0.303		  \\
							    & \emph{ Pipelining}   & 0.001			   & 0.056		  \\ \hline
\multirow{5}{*}{OuterProduct}   & \emph{Tile Size A}   & 0.095			   & \bf{0.290}	  \\ 
							    & \emph{Tile Size B}   & 0.075			   & \bf{0.323}	  \\ 
							    & \emph{OP}		       & 0.170			   & 0.084	      \\ 
							    & \emph{IP} 	       & \bf{0.321}		   & \bf{0.248}	  \\ 
							    & \emph{Pipelining}	   & \bf{0.340}	   	   & 0.055			  \\ \hline
\end{tabular}
\caption{Parameter feature importance per benchmark. Tile sizes are given for each data structure. OP and IP are the outer and inner loop parallelization factors, respectively. Pipelining determines whether the key compute loop in the benchmark is pipelined or sequentially executed. Scores closer to 1 mean that the parameter is more important for that objective. Scores for a single objective sum to 1.}
\label{table_feature_importance}
\end{table}

HyperMapper 2.0 can be used by domain non-experts to understand more about the domain they are trying to optimize. 
In particular, users can view feature importance to gain a better understanding of the impact of various parameters on the design objectives. 
The feature importances for the BlackScholes and OuterProduct benchmarks are given in Table~\ref{table_feature_importance}. 


In BlackScholes, innermost loop parallelization (IP) directly determines how fast a single tile of data can be processed. Consequently, as shown in Figure~\ref{table_feature_importance}, IP is highly related to both the design logic utilization and design run-time (cycles).
Since BlackScholes is bandwidth bound, changing DRAM utilization with tile sizes directly changes the run-time, but has no impact on the compute logic since larger memories do not require more LUTs. 
Outer loop parallelization (OP) also duplicates compute logic by making multiple copies of each inner loop, but as shown in Figure~\ref{table_feature_importance}, OP has less importance for run-time than IP.

Similarly, in OuterProduct, both tile sizes have roughly even importance on the number of execution cycles, while IP has roughly even importance for both logic utilization and cycles. Unlike BlackScholes, which includes a large amount of floating point compute, OuterProduct has relatively little computation, making the cost of outer loop pipelining relatively impactful on logic utilization but with little importance on cycles. In both cases, this information is taken into account when determining whether to prioritize further optimizing the application for inner loop parallelization or outer loop pipelining.

\subsection{Optimization Wall-clock Time}
\label{optimization_speed}
Since HyperMapper 1.0 was not optimized for tuning time \cite{nardi2017algorithmic}, our framework is already one order of magnitude faster in average. We run on four Intel XEON E7-8890 at 2.50GHz but HyperMapper 2.0 runs mostly sequentially. Optimization wall-clock time varies with the benchmark and is in the same order of magnitude as the Spatial baseline (tens of seconds). This includes the time to evaluate the Spatial samples which is independent from HyperMapper 2.0. 




\section{Related Work}
\label{related_work}
During the last two decades, several design space exploration techniques and frameworks 
have been used in a variety of different contexts ranging from embedded devices to compiler research to system integration.
Table~\ref{table_taxonomy_software} provides a taxonomy of methodologies and software from both the computer systems and machine learning communities. 
HyperMapper 2.0 has been inspired by a wide body of work in multiple sub-fields of these communities. 
The nature of computer systems workloads brings some important features to the design of HyperMapper 2.0 which are often missing in the machine learning community research on design space exploration tools.

In the system community, a popular, state-of-the-art design space exploration tool is OpenTuner \cite{ansel2014opentuner}.
This tool is based on direct approaches (\eg, differential evolution, Nelder-Mead) and a methodology based on the Area Under the Curve (AUC) and multi-armed bandit techniques to decide what search algorithm deserves to be allocated a higher resource budget. 
OpenTuner is different from our work in a number of ways. 
First, our work supports multi-objective optimization. Second, our white-box model-based approach enables the user to understand the results while learning from them. Third, our approach is able to consider unknown feasibility constraints. Lastly, our framework has the ability to inject prior knowledge into the search.  
The first point in particular does not allow a direct performance comparison of the two tools. 

Our work is inspired by HyperMapper 1.0~\cite{bodin2016integrating,nardi2017algorithmic,saeedi2017application,koeplinger2018}.
Bodin~\etal~\cite{bodin2016integrating} introduce HyperMapper 1.0 for autotuning of computer vision applications by considering the full software/hardware stack in the optimization process. 
Other prior work applies it to computer vision and robotics applications ~\cite{nardi2017algorithmic,saeedi2017application}. There has also been preliminary study of applying HyperMapper 1.0 to the Spatial programming language and compiler like in our work~\cite{koeplinger2018}.
However, it lacks fundamental features that makes it ineffective in the presence of applications with non-feasible designs and prior knowledge. 

In \cite{ipek2006efficiently} the authors use an active learning technique to build an accurate surrogate model by reducing the variance of an ensemble of fully connected neural network models. However, our work is fundamentally different because we are not interested in building a perfect surrogate model, instead we are interested in optimizing the surrogate model (over multiple objectives). So, in our case building a very accurate surrogate model over the entire space would result in a waste of samples. 

Recent work \cite{cianfriglia2018model} uses decision trees to automatically tune discrete NVIDIA and SoC ARM GPUs. 
Norbert \etal tackle the software configurability problem for binary~\cite{siegmund2012predicting} and 
for both binary and numeric options~\cite{siegmund2015performance} using a performance-influence model which is based on linear regression. 
They optimize for execution time on several examples exploring algorithmic and compiler spaces in isolation.




In particular, machine learning (ML) techniques have been recently employed in both architectural and compiler research. 
Khan \etal~\cite{khan2007using} employed predictive modeling for cross-program design space exploration in multi-core systems. 
The techniques developed managed to explore a large design space of chip-multiprocessors running parallel applications with low prediction error. 
In \cite{balaprakash2016automomml} Balaprakash \etal introduce AutoMOMML, an end-to-end, ML-based framework to build predictive models for objectives such as performance, and power. 
\cite{balaprakash2013active} presents the ab-dynaTree active learning parallel algorithm that builds surrogate performance models for scientific kernels and workloads on single-core, multi-core and multi-node architectures. 
In \cite{zuluaga2013active} the authors propose the Pareto  Active  Learning
(PAL) algorithm which intelligently samples the design space to predict the Pareto-optimal set.

Our work is similar in nature to the approaches adopted in the Bayesian optimization literature \cite{shahriari2016taking}.
Example of widely used mono-objective Bayesian DFO software are SMAC \cite{hutter2011sequential}, SpearMint \cite{snoek2012practical,snoek2015scalable} and the work on tree-structured Parzen estimator (TPE) \cite{bergstra2011algorithms}.
These mono-objective methodologies are based on random forests, Gaussian processes and TPEs making the choice of learned models varied. 

\section{Conclusions and Future Work}
\label{conclusions}
HyperMapper 2.0 is inspired by HyperMapper 1.0 \cite{bodin2016integrating}, by the philosophy behind OpenTuner \cite{ansel2014opentuner} and SMAC \cite{hutter2011sequential}. 
We have introduced a new DFO methodology and corresponding framework which uses guided search using active learning. This framework, dubbed HyperMapper 2.0, is built for practical, user-friendly design space exploration in computer systems, including support for categorical and ordinal variables, design feasibility constraints, multi-objective optimization, and user input on variable priors. Additionally, HyperMapper 2.0 uses randomized decision forests to model the searched space. This model not only maps well for the discontinuous, non-linear spaces in computer systems, but also gives a ``white box'' result which the end user can inspect to gain deeper. 

We have presented the application of HyperMapper 2.0 as a compiler pass of the Spatial language and compiler for generating application accelerators on FPGAs. 
Our experiments show that, compared to the previously used heuristic random search, our framework finds similar or better approximations of the true Pareto frontier, with significantly fewer samples required, 8x in most of the  benchmarks explored. 

Future work will include analysis and incorporation of other DFO strategies. 
In particular, the use of a full Bayesian approach will help to leverage the prior knowledge by computing a posterior distribution. 
In our current approach we only exploit the prior distribution at the level of the initial warm-up sampling. 
Exploration of additional methods to warm-up the search from the design of experiments literature is a promising research venue. 
In particular the Latin Hypercube sampling technique was recently adapted to work on categorical variables \cite{swiler2014surrogate} making it suitable for computer systems workloads. 
Future work will target an extension of this work to the Spatial ASIC back end as well as the Halide programming language.

\section{Acknowledgements}
The authors wish to thank the HyperMapper 2.0 users for their help in improving the framework. 
We thank Professor Joseph Salmon for feedback on the manuscript and Matthew Feldman for compiler support. 
This research is supported by affiliate members and
supporters of the Stanford DAWN project: Ant Financial,
Facebook, Google, Infosys, Intel, Microsoft, NEC, Teradata,
SAP, and VMware.
\bibliographystyle{plain}
\bibliography{references}

\end{document}